\documentclass[preprint,12pt,authoryear]{elsarticle}

\usepackage{amssymb}
\usepackage{amsmath}
\usepackage{graphicx}
\usepackage{dcolumn}
\usepackage{subcaption}
\graphicspath{{./figures/}}
\usepackage{bm}
\usepackage{hyperref} 
\usepackage{natbib}
\usepackage{url}
\usepackage{tikz}
\usepackage{booktabs}
\usepackage{multirow}
\usepackage{array}
\usepackage{tabularx}
\usepackage{todonotes}
\usetikzlibrary{positioning, shapes.geometric, arrows.meta, calc}

\journal{Neuroscience Informatics}

\begin{document}

\begin{frontmatter}

\title{TRELLIS-Enhanced Surface Features for Comprehensive Intracranial Aneurysm Analysis}
\author{Clément Hervé}
\ead{clement.herve@etu.minesparis.psl.eu}
\author{Paul Garnier}
\author{Jonathan Viquerat}
\author{Elie Hachem}
\affiliation{organization={Mines Paris, PSL University, Centre for Material Forming (CEMEF), UMR CNRS 7635},
            addressline={rue Claude Daunesse}, 
            city={Sophia-Antipolis},
            postcode={06904}, 
            state={Alpes-Maritimes},
            country={France}}

\begin{abstract}
Intracranial aneurysms pose a significant clinical risk yet are difficult to detect, delineate, and model due to limited annotated 3D data. We propose a cross-domain feature-transfer approach that leverages the latent geometric embeddings learned by TRELLIS, a generative model trained on large-scale non-medical 3D datasets, to augment neural networks for aneurysm analysis. By replacing conventional point normals or mesh descriptors with TRELLIS surface features, we systematically enhance three downstream tasks: (i) classifying aneurysms versus healthy vessels in the Intra3D dataset, (ii) segmenting aneurysm and vessel regions on 3D meshes, and (iii) predicting time-evolving blood-flow fields using a graph neural network on the AnXplore dataset. Our experiments show that the inclusion of these features yields strong gains in accuracy, F1-score, and segmentation quality over state-of-the-art baselines, and reduces simulation error by 15\%. These results illustrate the broader potential of transferring 3D representations from general-purpose generative models to specialized medical tasks.\end{abstract}

\end{frontmatter}

\section{Introduction} \label{INTRO}

\begin{figure}[!h]
  \centering
  \includegraphics[width=0.55\textwidth]{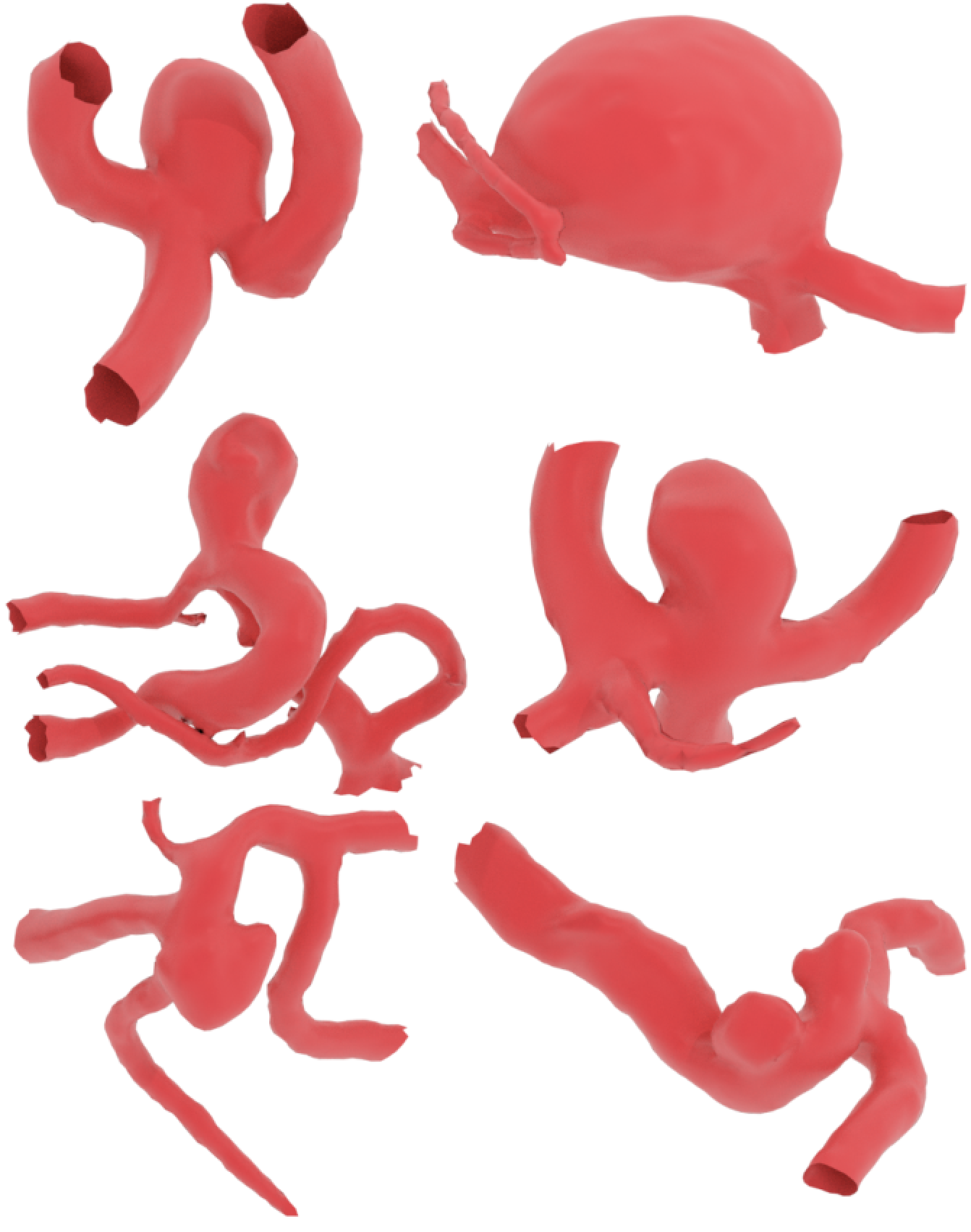}
  \caption{Examples of intracranial aneurysms and vessels from the Intra3D dataset.}
  \label{fig:intra_aneurysms}
\end{figure}

\begin{figure}[!h]
  \centering
  \includegraphics[width=1\textwidth]{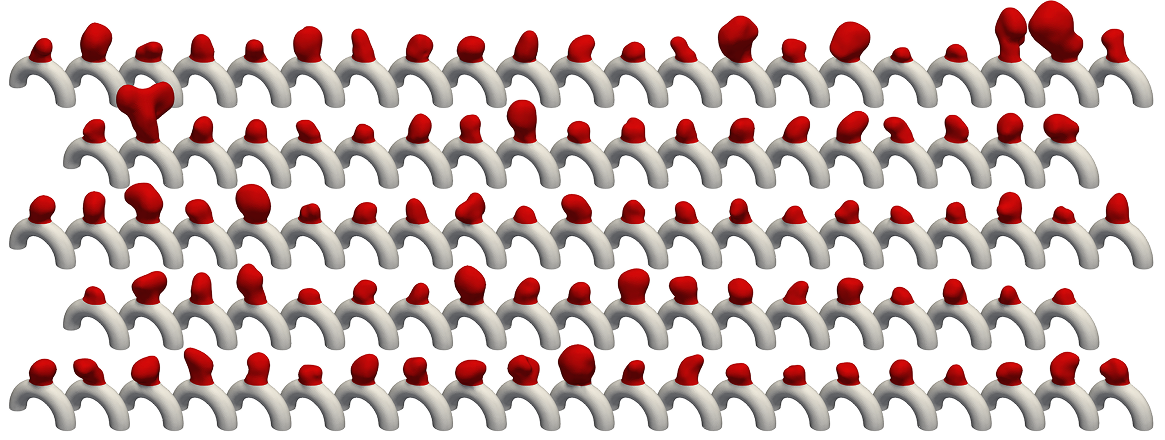}
  \caption{Examples of aneurysms from the AnXplore dataset}
  \label{fig:anxplore_aneurysms}
\end{figure}

Intracranial aneurysms represent a significant and often silent threat to human health. These pathologies typically remain asymptomatic until the moment of rupture, a catastrophic event associated with high rates of morbidity and mortality. Consequently, the early and accurate detection of unruptured aneurysms is crucial for preventive clinical intervention. The primary challenge in their identification lies in their insidious nature; discovery is frequently incidental, occurring during diagnostic procedures for unrelated conditions. Advanced imaging modalities, particularly Magnetic Resonance Angiography (MRA), have become instrumental in this endeavor, enabling the reconstruction of detailed 3D models of the brain's vascular network to facilitate the identification of aneurysms.
To advance research in this area, publicly available datasets are indispensable. The Intra3D dataset \citep{yang2020intra} (see \autoref{fig:intra_aneurysms}), a component of the larger MedMNISTv2 set \citep{Yang_2023}, provides a valuable resource of 3D vascular models, including both healthy vessels and aneurysms, in various representations such as meshes and point clouds. The original study introducing this dataset established baseline performance for classification and segmentation tasks using neural networks \citep{yang2020intra}. Subsequent research has seen the development of more sophisticated models, including 3DMedPT \citep{yu20213dmedicalpointtransformer} and GRAB-net \citep{10093984}, which have demonstrated state-of-the-art performance on this benchmark.
Beyond detection, the management of diagnosed aneurysms often necessitates surgical intervention to mitigate the risk of rupture. The planning of such complex procedures is increasingly supported by computational fluid dynamics (CFD) simulations, which offer invaluable insights into patient-specific hemodynamics. The advent of physics-informed neural networks \citep{Arzani_2021} has marked a significant milestone, paving the way for mesh-based neural networks that can accurately model hemodynamic phenomena \citep{Suk_2024, graphphysics}. These AI-driven methods can predict intricate blood flow patterns and their interactions with vessel walls, thereby enriching the information available for clinical decision-making.
In this context, the AnXplore dataset \citep{anxplore} (see \autoref{fig:anxplore_aneurysms}) was developed, comprising 101 aneurysm models from the Intra3D dataset \citep{yang2020intra} paired with their corresponding blood flow simulations. This dataset is particularly well-suited for training graph neural networks (GNNs) \citep{pfaff2021learningmeshbasedsimulationgraph} to predict hemodynamic patterns \citep{graphphysics}. The architectural foundation for such models is often an encode-process-decode framework, which, when augmented with attention mechanisms \citep{VaswaniSPUJGKP17}, can achieve strong simulation fidelity.

Concurrently, the field of generative artificial intelligence has witnessed extraordinary progress in 3D object synthesis. A notable example is TRELLIS \citep{xiang2024structured}, an algorithm capable of generating high-fidelity 3D objects from various inputs, including text, 2D images, and 3D assets. Its utility extends to engineering and physics applications, such as motion planning with IMPACT \citep{ling2025impactintelligentmotionplanning} and deformable object reconstruction with PhysTwin \citep{jiang2025phystwinphysicsinformedreconstructionsimulation}. The model was trained on extensive and diverse datasets like Objaverse (XL) \citep{deitke2023objaversexluniverse10m3d}, ABO \citep{collins2022abodatasetbenchmarksrealworld}, 3DFUTURE \citep{fu20203dfuture3dfurnitureshape}, and HSSD \citep{khanna2023habitatsyntheticscenesdataset}. However, these foundational datasets lack specific medical geometries, such as cerebral aneurysms and blood vessels.
This work bridges the gap between these distinct but complementary fields. We hypothesize that the rich, latent representations learned by a state-of-the-art generative model, such as TRELLIS, can significantly enhance the performance of discriminative models in a specialized medical domain. To test this, we leverage the TRELLIS encoder to extract deep geometric features from the Intra3D \citep{yang2020intra} and AnXplore \citep{anxplore} datasets. These features are then integrated into established neural network architectures for analyzing aneurysms. Specifically, we augment PointNet \citep{pointnet} and PointNet++ \citep{pointnetpp} for classification and segmentation tasks and enhance a mesh-based GNN for blood flow simulation as well. Our central objective is to demonstrate that these TRELLIS-derived features can provide a more potent signal for these networks, leading to improved accuracy and predictive power.
Our methodology involves mapping the 3D vascular objects into the latent space of the TRELLIS encoder to obtain compact embeddings and point representations. These features are then used as supplementary input to the downstream neural networks. We demonstrate that this feature enhancement strategy yields substantial improvements in both aneurysm classification and segmentation, as well as in the fidelity of hemodynamic simulations. Furthermore, we conduct a thorough investigation into the nature of these learned features. We conduct an ablation study to evaluate the effectiveness of using these features alone for classification. Through Principal Component Analysis (PCA) and t-distributed Stochastic Neighbor Embedding (t-SNE), we visualize the feature space to gain insight into its underlying structure. We also apply clustering algorithms to the AnXplore dataset to explore correlations between latent space proximity and geometric similarity. The code for this research is publicly available at \url{https://github.com/clementhrv/trellis_for_intra}.

The remainder of this paper is structured as follows. \autoref{DATASET} provides a detailed description of the datasets employed in our study. \autoref{TRELLIS} elaborates on the TRELLIS encoding process and presents our analysis of the extracted features. In \autoref{POINTCLOUD}, we detail the point cloud-based methods for classification and segmentation. \autoref{GNN} introduces the graph neural network architecture for blood flow simulation. \autoref{RESULTS} presents and discusses our experimental findings. Finally, \autoref{CONCLUSION} summarizes our contributions and outlines potential avenues for future research.

\section{Datasets} \label{DATASET}

\subsection{Intra3D Dataset} \label{INTRA3D}

The Intra3D dataset \citep{yang2020intra} comprises 3D models of intracranial aneurysms and healthy blood vessels, reconstructed from 2D MRA scans, for classification and segmentation tasks. The classification dataset is imbalanced, comprising 1,694 healthy vessels and 215 aneurysms, which can be combined with the 116 annotated aneurysms used to train segmentation models, resulting in a total of 331 aneurysms for classification. The original paper also presents classification and segmentation results using models such as PointNet \citep{pointnet}, PointNet++ \citep{pointnetpp}, PointCNN \citep{pointcnn}, SO-net \citep{sonet}, and PointConv \citep{pointconv}.

\subsection{AnXplore Dataset} \label{ANXPLORE}

The AnXplore dataset \citep{anxplore} comprises 101 3D models of intracranial aneurysms, each accompanied by associated blood flow simulations in the form of volumetric data. These aneurysms are extracted from the 116 annotated aneurysms in Intra3D \citep{yang2020intra}. To generate each model, the head of each aneurysm was isolated and placed on the same uniform vessel, meaning all aneurysms in AnXplore \citep{anxplore} are located on the same vessel but vary in shape and size. 

\section{TRELLIS} \label{TRELLIS}

\subsection{TRELLIS encoding}

\begin{figure*}
  \centering
  \includegraphics[width=\textwidth]{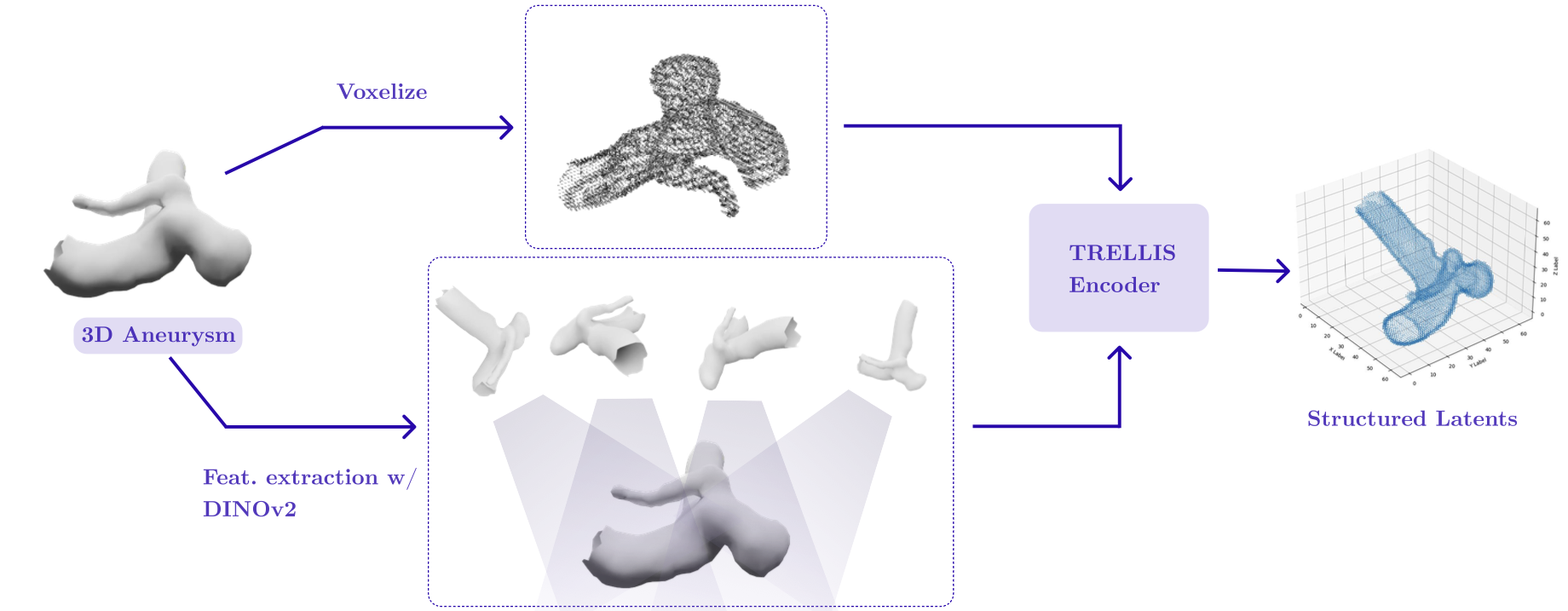}
  \caption{Illustrations of the entire pipeline based on TRELLIS to extract features for each aneurysm.}
  \label{fig:vae}
\end{figure*}

\noindent TRELLIS \citep{xiang2024structured} is an advanced algorithm designed to generate 3D objects from diverse inputs, such as text prompts, multiple 2D views of an object, or another 3D object directly. For the 3D object-to-3D object transformation, TRELLIS employs a robust encoding process to extract detailed features from a mesh, facilitating accurate object reconstruction. This encoding process involves several key steps:
first, the 3D object is rendered from multiple angles to capture its geometric details, and then, these rendered views are voxelized, converting the object into a three-dimensional grid of pixels (voxels) that represent its structure. In our case, this grid is a $64 \times 64 \times 64$ array with approximately 5,000 active voxels representing the object's surface.
The voxelized object is then processed to calculate features for each voxel. TRELLIS uses the voxelized representation combined with random views of the object, which are processed using a pre-trained DINOv2 autoencoder \citep{oquab2024dinov2learningrobustvisual}. It is based on a vision transformer architecture (ViT) \citep{dosovitskiy2021imageworth16x16words}, and extracts high-dimensional features from the rendered images. It is trained to learn robust visual representations and used to encode each voxel in the voxelized grid.
The resulting feature-position pairs are passed into a transformer-based sparse variational autoencoder (VAE). The VAE serializes the inputs with positional encodings and processes them using shifted window attention to model local interactions efficiently. The encoder outputs structured latent tokens, each represented by a 1024-dimensional vector associated with a specific voxel position.
The TRELLIS encoder has already been trained using 500,000 3D assets from 4 public datasets, Objaverse (XL) \citep{deitke2023objaversexluniverse10m3d} , ABO \citep{collins2022abodatasetbenchmarksrealworld}, 3DFUTURE \citep{fu20203dfuture3dfurnitureshape}, and HSSD \citep{khanna2023habitatsyntheticscenesdataset}. We used the encoder directly on Intra3D \citep{yang2020intra} and AnXplore \citep{anxplore} datasets to extract the features from the meshes of aneurysms and vessels.
This process is illustrated in \autoref{fig:vae}.

\subsection{Analysis of the TRELLIS features} 

\begin{figure}[h]
  \centering
  \includegraphics[width=0.32\textwidth]{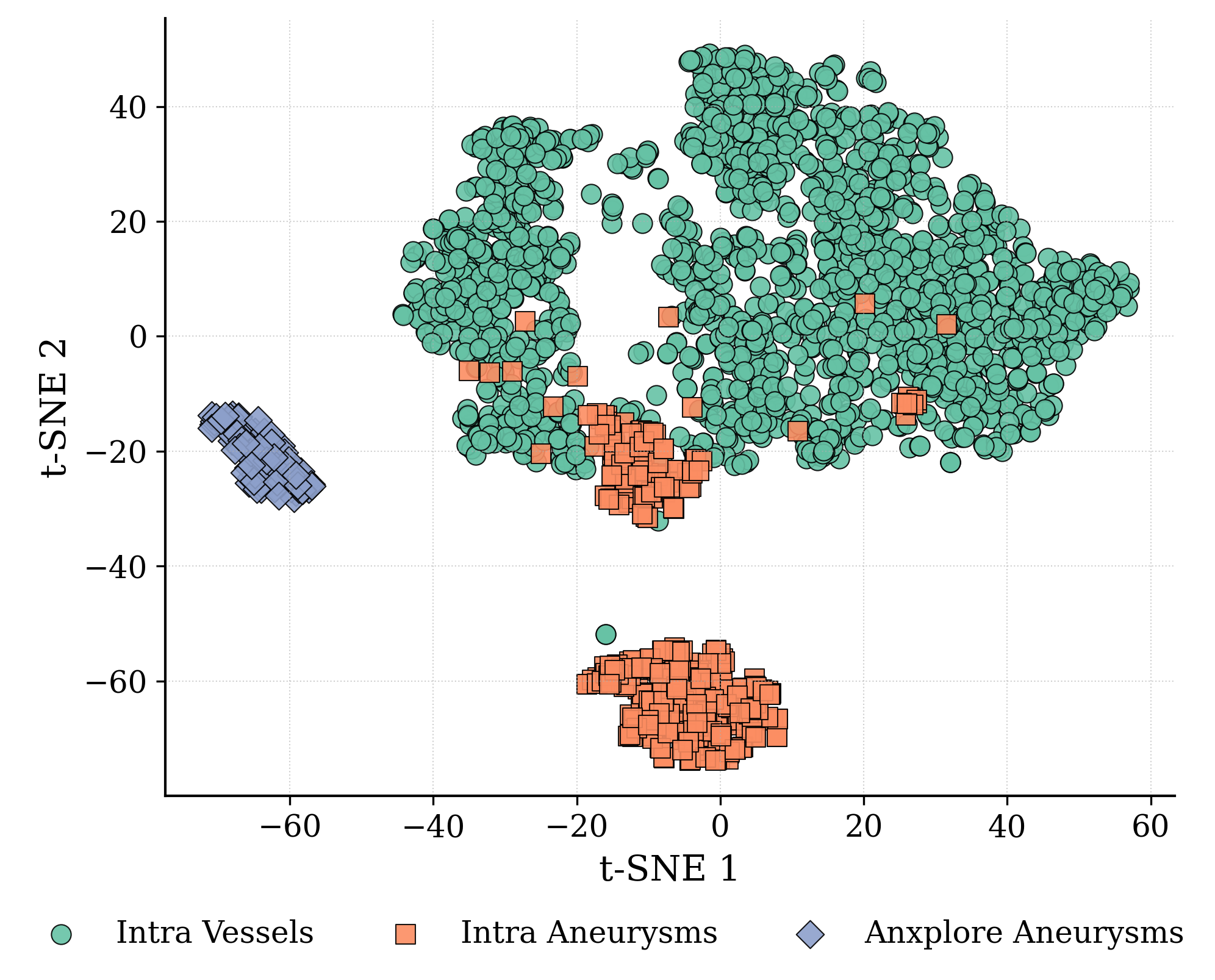}
  \includegraphics[width=0.32\textwidth]{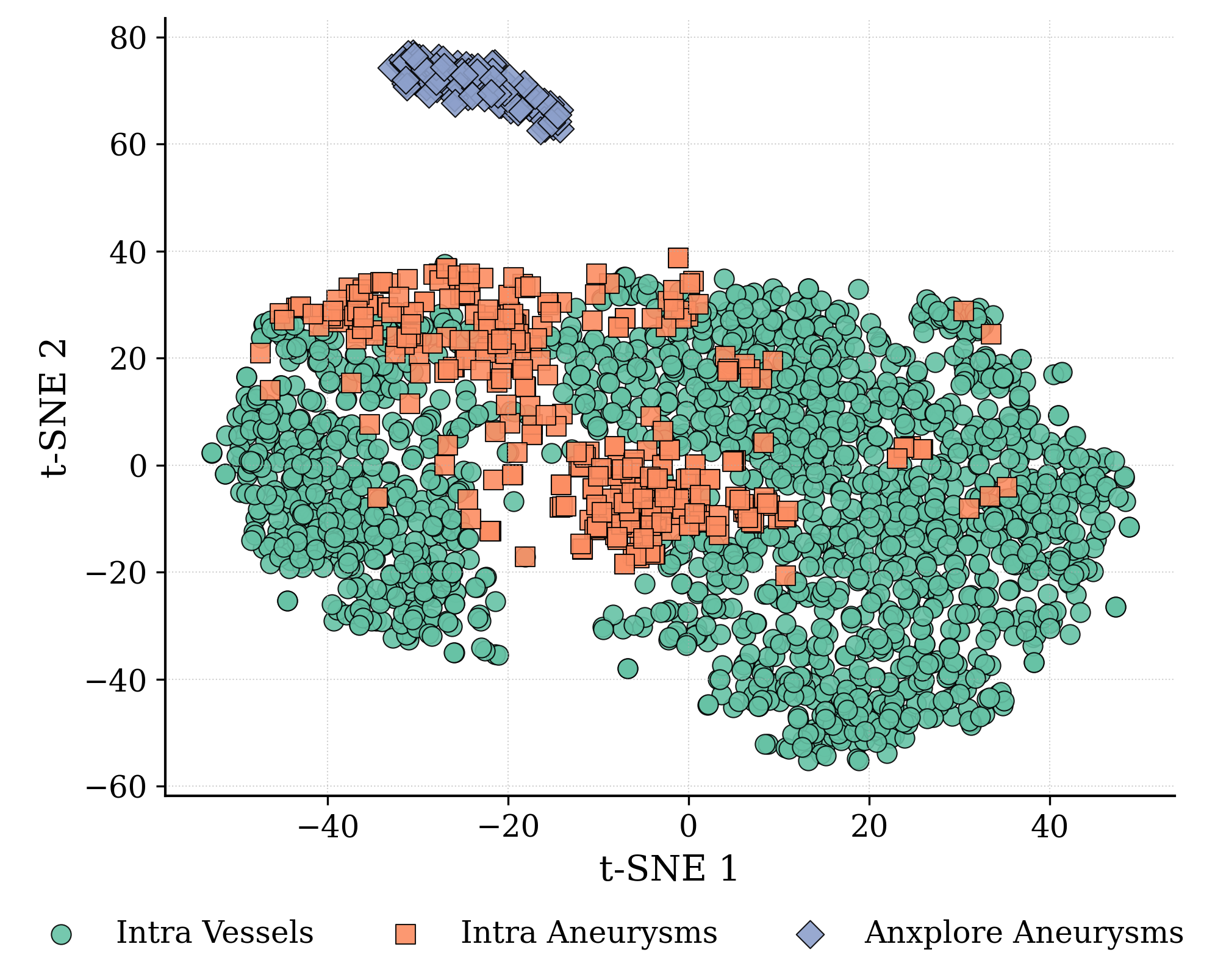}
  \includegraphics[width=0.32\textwidth]{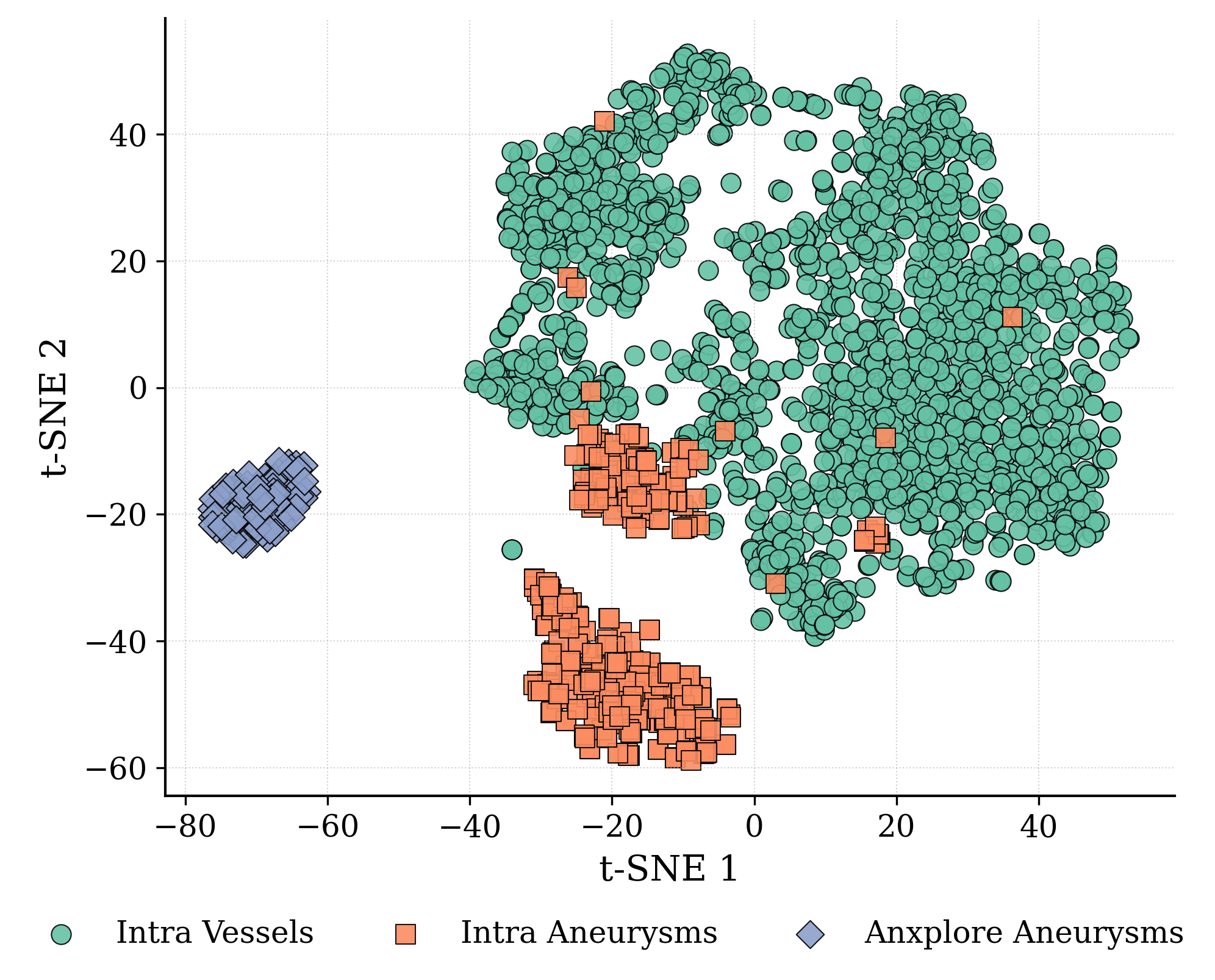}
  \caption{Result of t-SNE on the Intra3D dataset for classification and AnXplore dataset. The left figure shows the mean features, the middle figure shows the standard deviation features, and the right figure shows the minimum features.}
  \label{fig:tsne}
\end{figure}

\noindent We conducted several analyses on the TRELLIS features to understand them better and to evaluate how well these features represent the objects.
First, for each object, we computed various statistical metrics. Since a 1024-dimensional feature vector describes each point, we calculated the mean, standard deviation, minimum, and maximum for each feature across all points in an object, synthesizing them into four 1024-dimensional vectors that characterize each object. We then performed 2D PCA on these vectors for each category, first on the Intra3D dataset \citep{yang2020intra}, and then on the combined Intra3D and AnXplore datasets \citep{anxplore}. The goal of the combined analysis was to assess whether the AnXplore aneurysms are encoded differently compared to the Intra3D aneurysms. Results for the mean and standard deviation features are shown in Figure \ref{fig:pca}; additional results are provided in the appendix.
To investigate whether other components could provide a clearer separation of aneurysms, we also performed t-SNE on the combined dataset, as shown in Figure \ref{fig:tsne} for the mean features. This technique is a dimensionality reduction technique that visualizes high-dimensional data by mapping it into a lower-dimensional space while preserving local similarities.

\begin{figure*}
  \centering
  \includegraphics[width=0.32\textwidth]{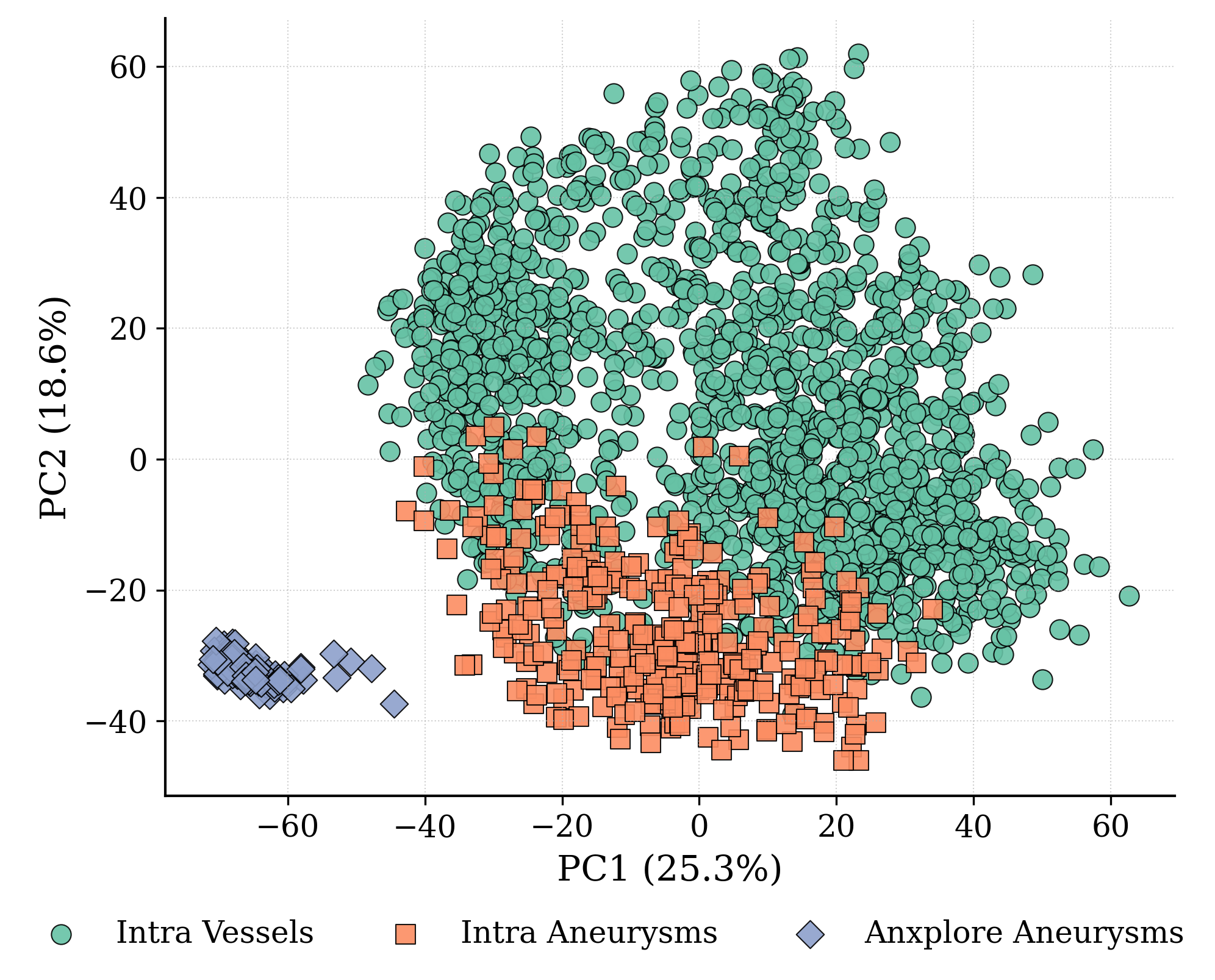}
  \includegraphics[width=0.32\textwidth]{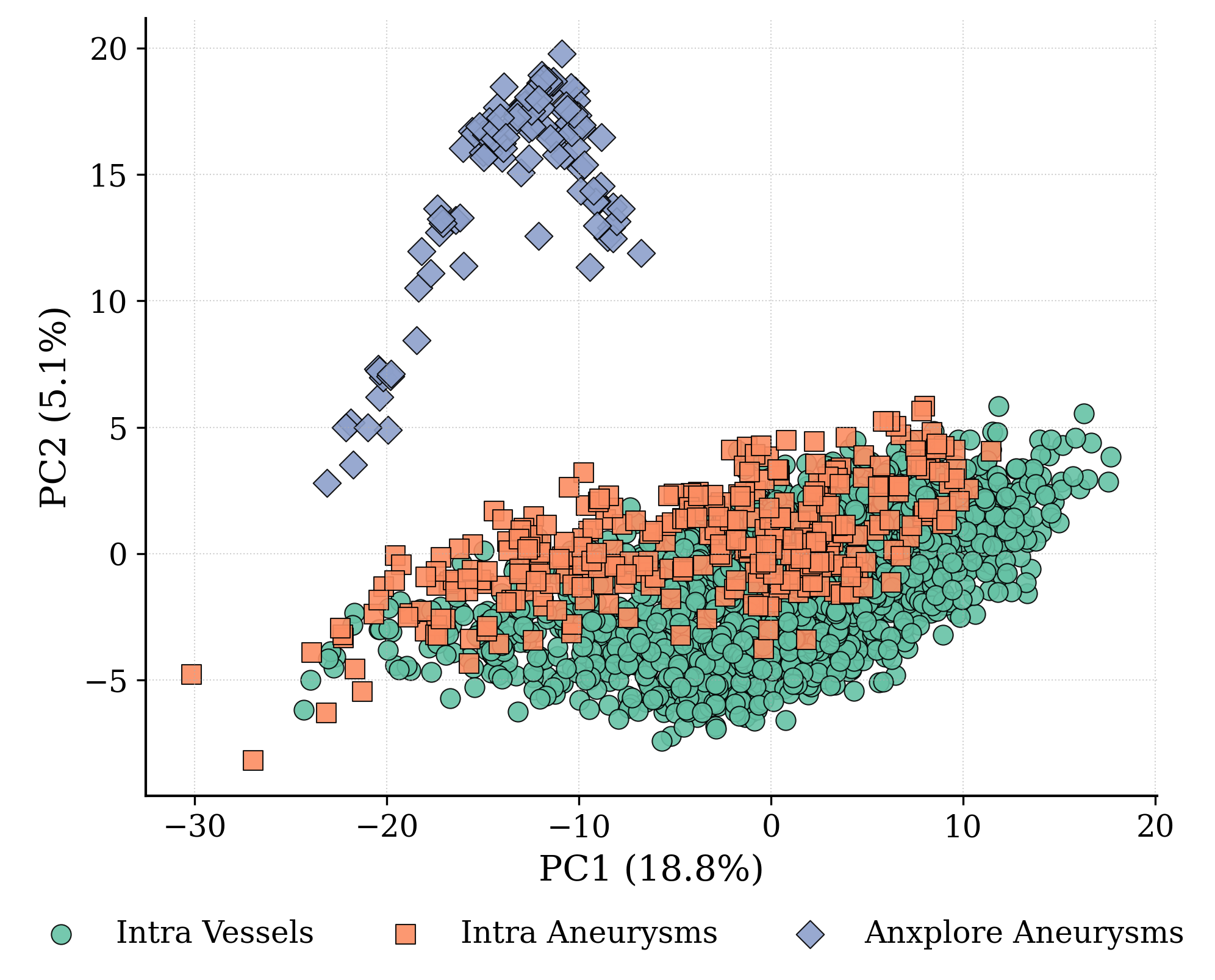}
  \includegraphics[width=0.32\textwidth]{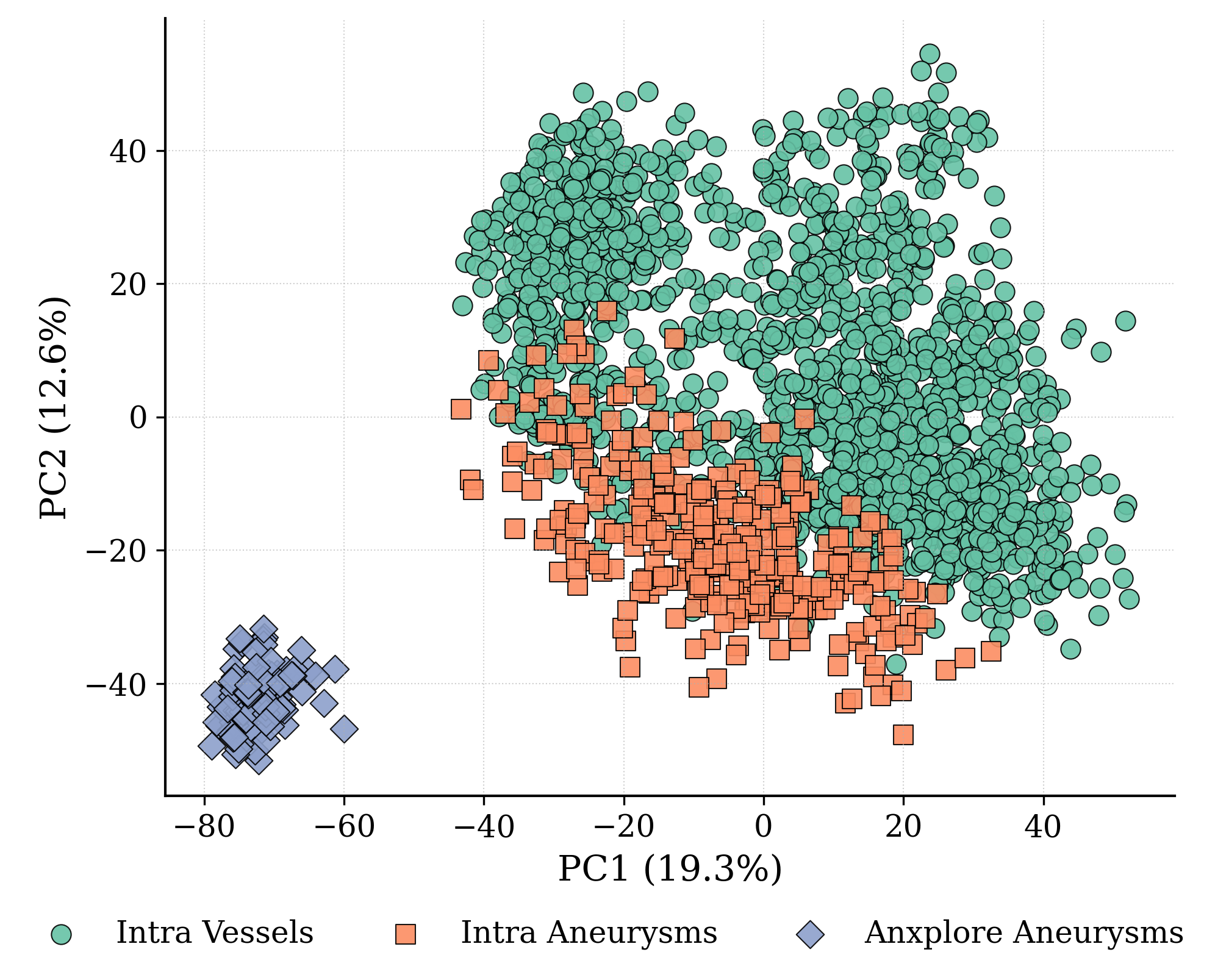}
  \caption{Results of PCA on the Intra3D dataset for classification and AnXplore dataset. The left figure shows the mean features, the middle figure shows the standard deviation features, and the right figure shows the minimum features.}
  \label{fig:pca}
\end{figure*}

\begin{figure}[h!]
  \centering
  \includegraphics[width=0.48\textwidth]{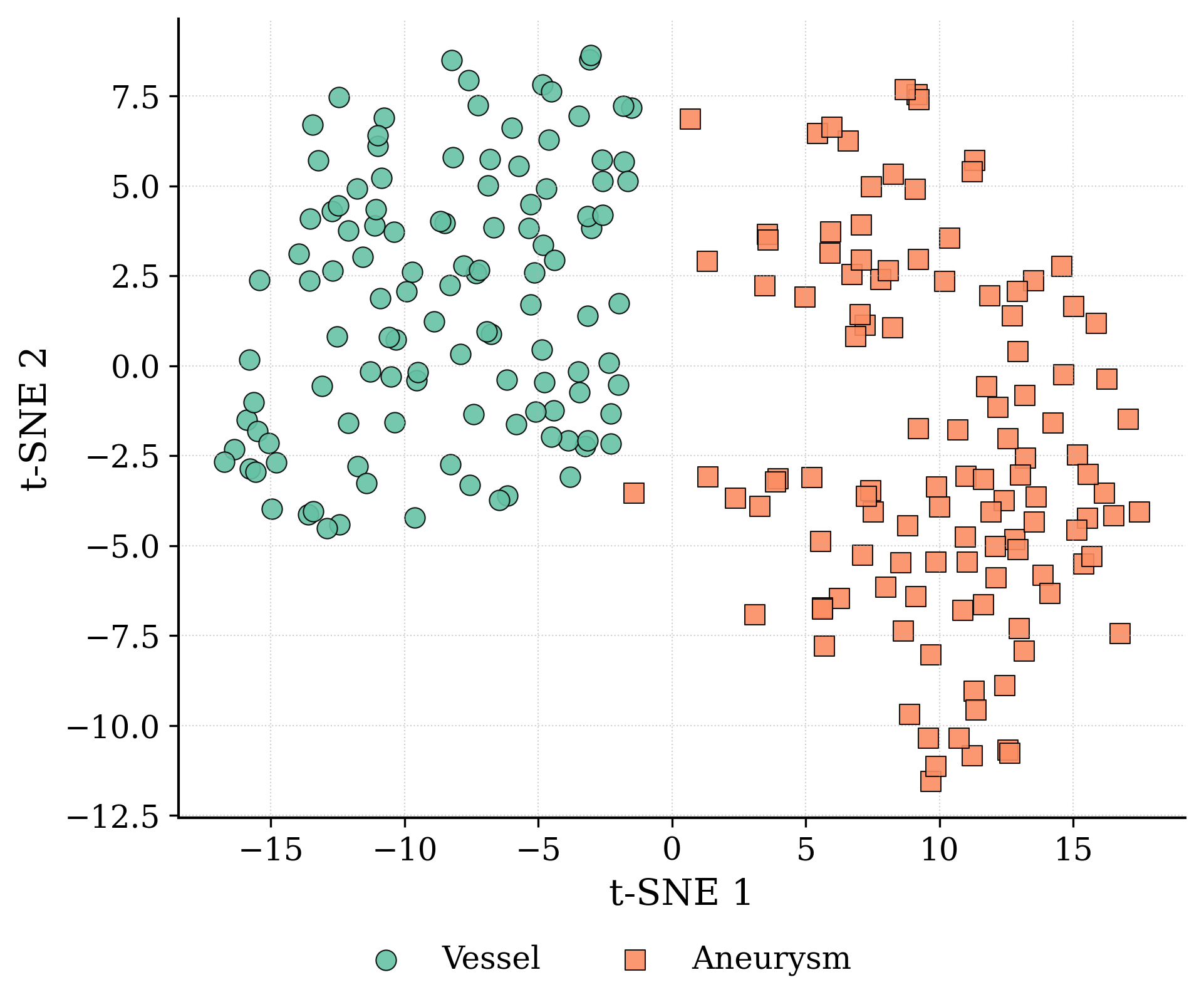}
  \includegraphics[width=0.48\textwidth]{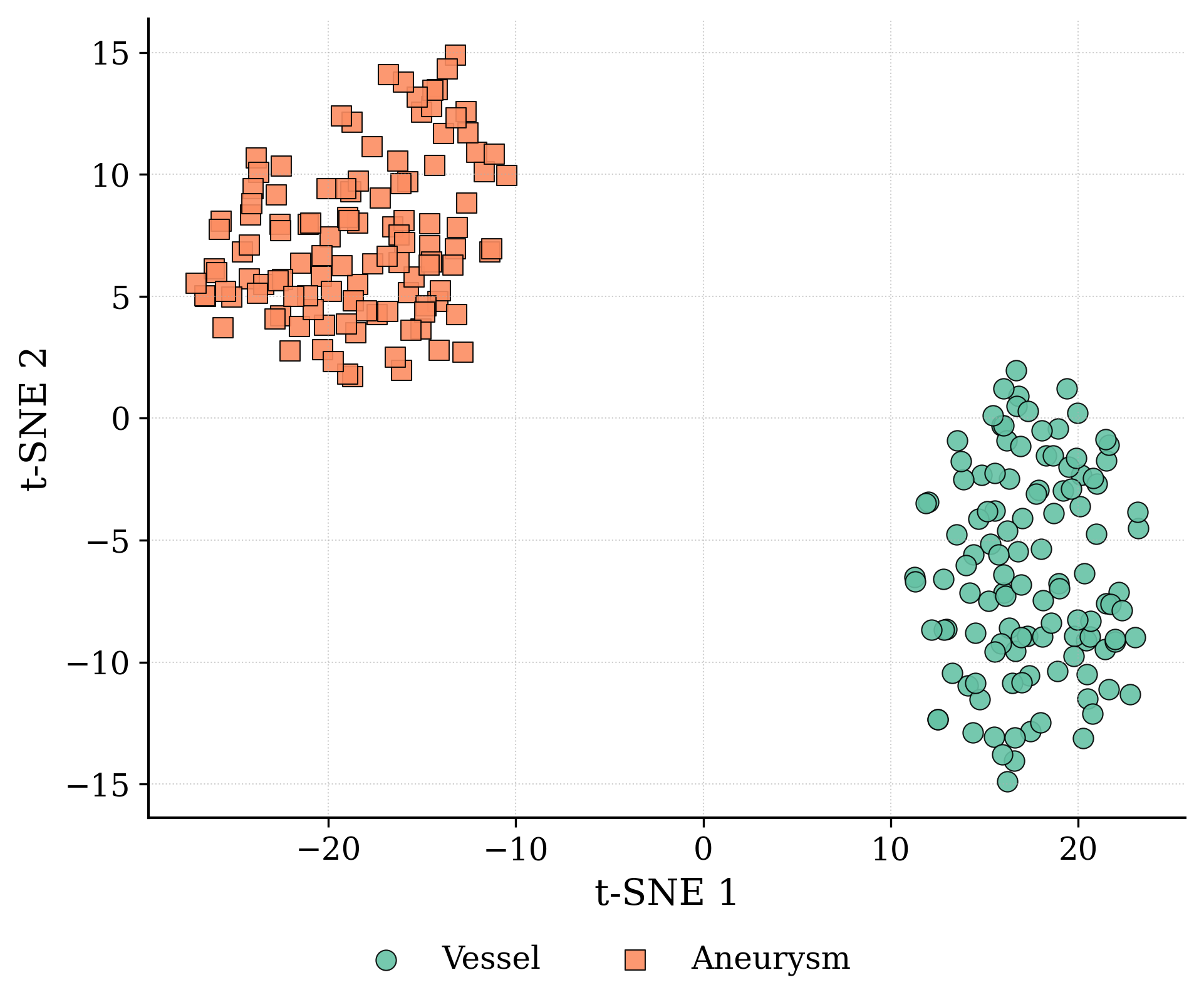}
  \caption{Results of t-SNE on annotated aneurysms from Intra3D \citep{yang2020intra} using the mean features of the aneurysm and vessel parts. The left figure shows the mean features, and the right figure shows the standard deviation features.}
  \label{fig:aneu_seg}
\end{figure}

\noindent For the Intra3D dataset alone, we observe that all the metrics related to the features separate aneurysms and vessels.  
When combining with the AnXplore dataset, we observe that the two datasets are separated for all four metrics, suggesting that the characteristics of the vessel on which an aneurysm is located have a strong influence on the extracted features. With t-SNE, the separation becomes even more pronounced: AnXplore is consistently well-separated from Intra3D, and within Intra3D, aneurysms and vessels are separated using the mean, maximum, and minimum features.
To further assess the utility of PCA and the extracted features for classification, we trained machine learning algorithms on the resulting 2D projections from the Intra3D dataset, as discussed in the following section.
For the segmentation dataset, we compute the mean, standard deviation, minimum, and maximum features for points from the aneurysm and vessel parts separately and perform t-SNE over the 116 annotated aneurysms. This proved to be highly effective and revealed that the two parts, the aneurysm and the vessel, are encoded distinctly. Results are shown in Figure~\ref{fig:aneu_seg}. To further study these differences, we also performed 2D PCA and t-SNE on all the features from a single element, which yielded a clear separation between the aneurysm and vessel parts. This indicates that the features extracted by TRELLIS \citep{xiang2024structured} are effective for distinguishing between these two components of the 3D model, as shown in Figure \ref{fig:aneu_seg_2}. All the figures can be found in the appendix.

\begin{figure}[h!]
  \centering
  \includegraphics[width=0.48\textwidth]{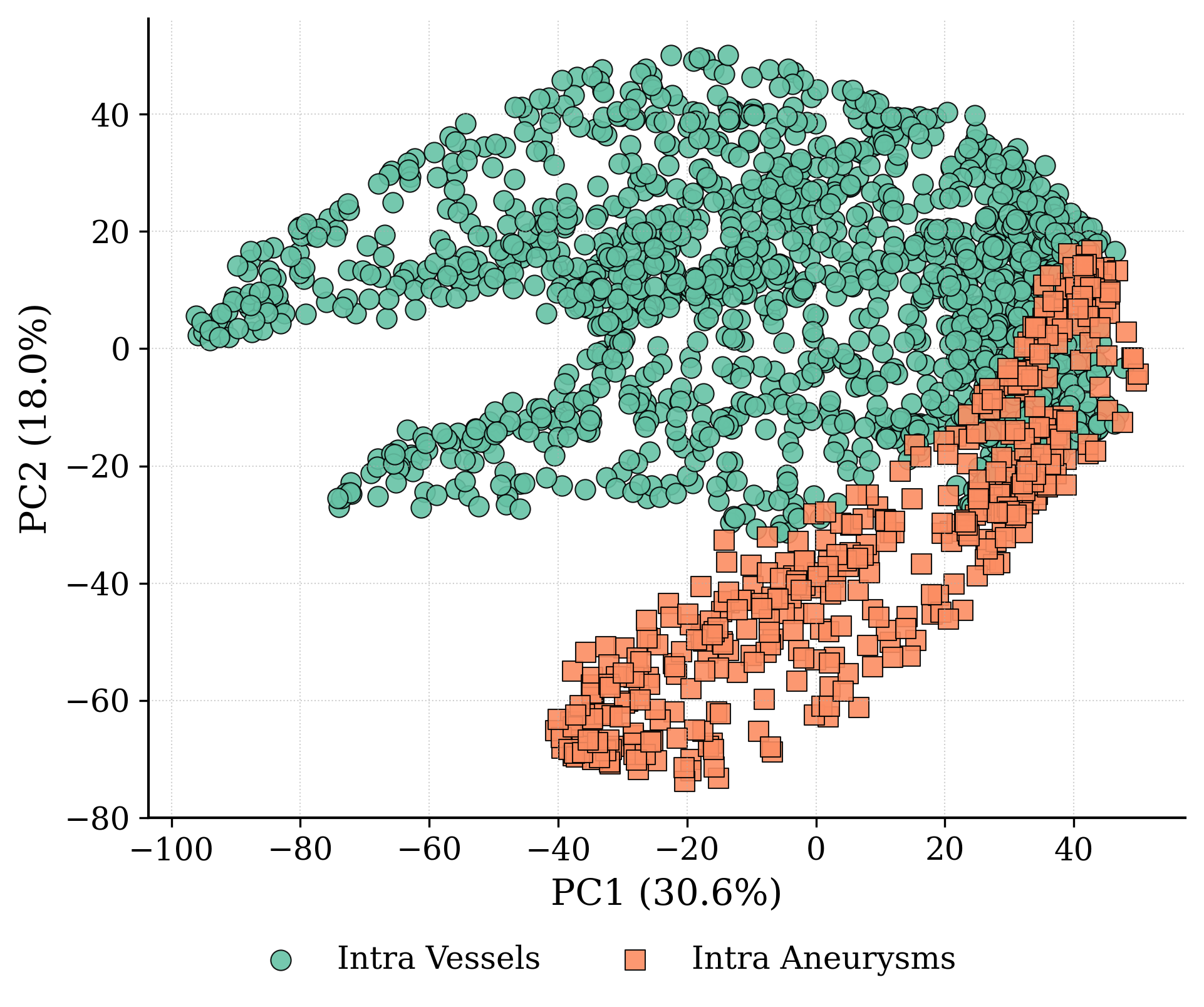}
  \includegraphics[width=0.48\textwidth]{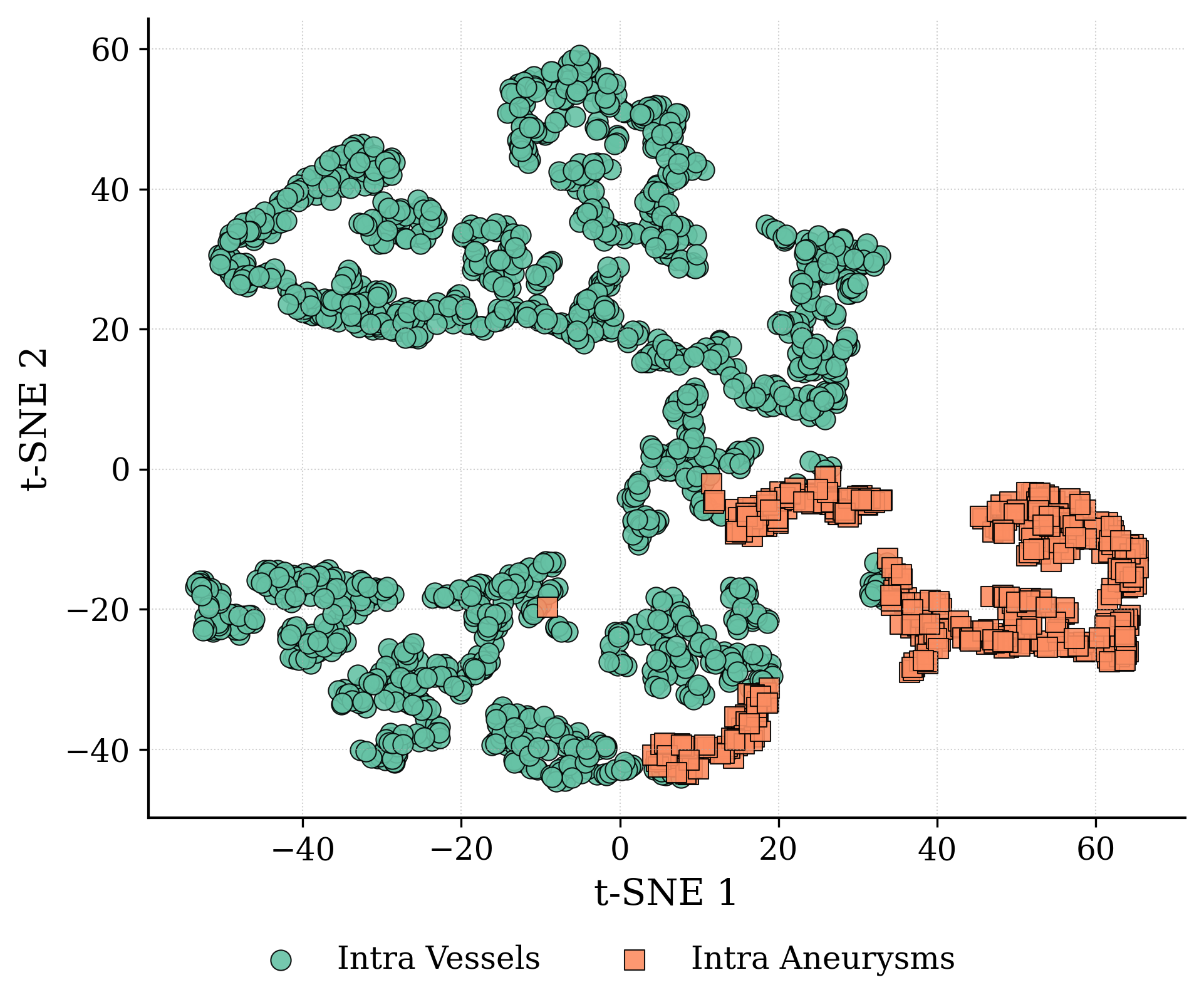}
  \caption{Results of PCA and t-SNE on all the features from an annotated aneurysm from Intra3D \citep{yang2020intra}. The left figure shows the PCA results, and the right figure shows the t-SNE results.}
  \label{fig:aneu_seg_2}
\end{figure}

\begin{figure}[h!]
  \centering
  \includegraphics[width=0.20\textwidth]{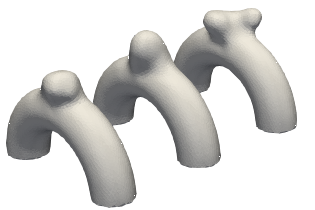}
  \includegraphics[width=0.27\textwidth]{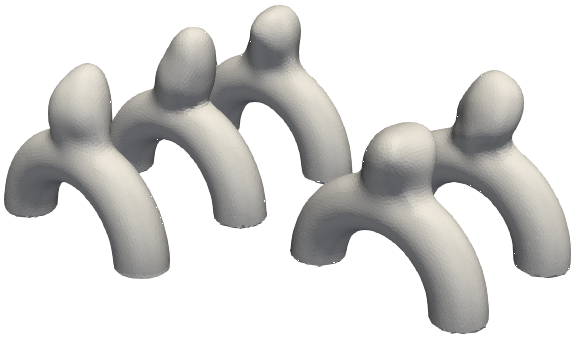}
  \includegraphics[width=0.28\textwidth]{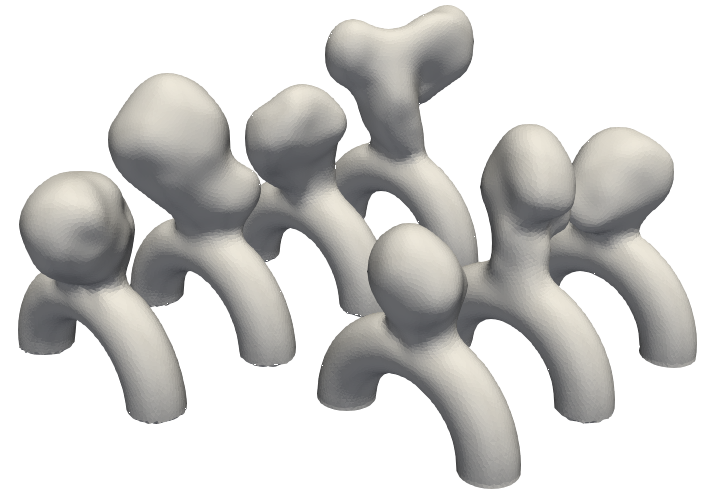}
  \includegraphics[width=0.18\textwidth]{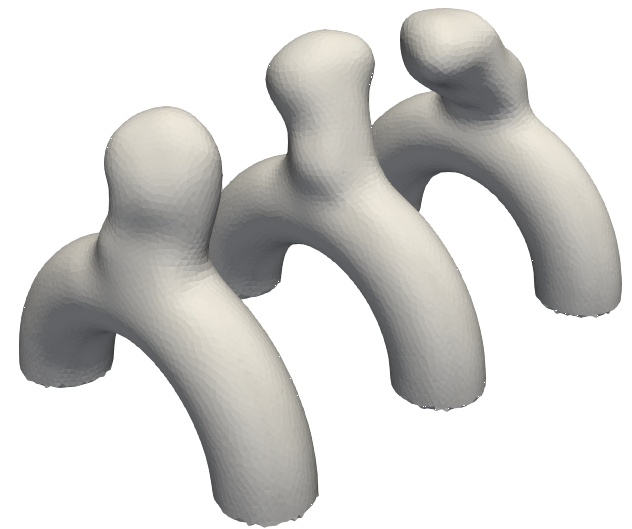}
  \caption{Results of clustering on the AnXplore dataset using PCA on the mean features. The figure shows four clusters of aneurysms.}
  \label{fig:clustering}
\end{figure}

\begin{figure}[h!]
  \centering
  \includegraphics[width=0.7\textwidth]{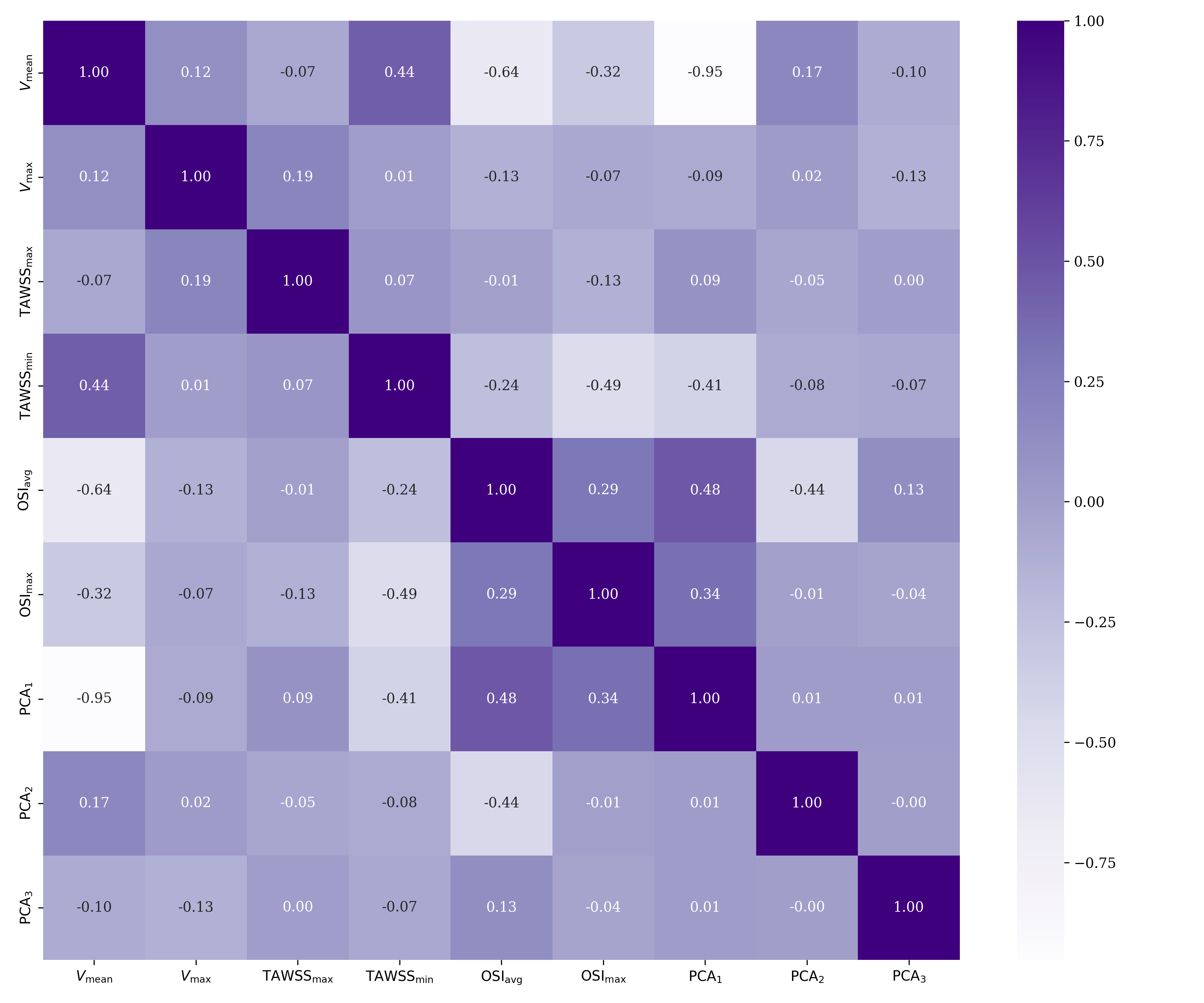}
  \caption{Correlation heatmap between the 3 main features from TRELLIS and several hemodynamic metrics on the AnXplore dataset. Metrics are computed following the same method as \cite{anxplore}.}
  \label{fig:heatmapcluster}
\end{figure}

Finally, we performed 2D PCA over the mean, standard deviation, minimum, and maximum of the features from each element of the AnXplore dataset and applied clustering algorithms to determine whether aneurysms within the same cluster exhibited similarities in shape and size. The optimal number of clusters for the 101 aneurysms was 15, and the results were revealing: within each cluster, aneurysms were generally similar in size and morphology, indicating that TRELLIS features are highly effective in capturing the geometric characteristics of 3D aneurysms. This suggests that, since aneurysms at higher risk of rupture tend to be larger and more irregular, the clustering output could be used to identify high-risk aneurysms. Results for the mean-based clusters are presented in Figure \ref{fig:clustering}. 
We further extend our analysis and study whether the same TRELLIS features can be used to predict more intricate metrics than geometry, such as hemodynamic metrics like the Time Averaged Wall-Shear-Stress (TAWSS) (see \autoref{fig:heatmapcluster}). 
While those features are obviously not enough for such a task, they are relevant and pave the way for a strategy where features from a non-medical foundational model could enhance or even replace tailored geometrical metrics (such as the bulge angle, the aneurysm diameter, or its non-sphericity index).

This analysis shows that the TRELLIS features are very effective in capturing the geometric characteristics of the objects, even medical data on which it has not been trained, enabling the classification and segmentation of aneurysms and vessels. The clustering results also indicate that the features can be used to identify aneurysms with similar shapes, which could be useful for risk assessment. Overall, this precise description shows that even without complex neural networks, the features could alone be used to detect aneurysms, and, combined with complex algorithms, could certainly improve the performance of classification and segmentation tasks.

\section{Point Cloud-Based Methods} \label{POINTCLOUD}

\subsection{Algorithms}

Point cloud-based methods are commonly used for 3D object classification and segmentation. We explored two models: PointNet \citep{pointnet} and PointNet++ \citep{pointnetpp} that are effective for processing 3D datasets, for example, in the Intra3D dataset \citep{yang2020intra} or for non-medical data such as ModelNet40 \citep{7298801}.
PointNet is a pioneering architecture that treats a point cloud \( \mathcal{X} = \{x_1, \dots, x_N\} \subset \mathbb{R}^3 \) as an unordered set of points. To maintain permutation invariance, the network applies a shared multi-layer perceptron (MLP) to each point independently and aggregates the resulting features using a symmetric function, typically max pooling:

\vspace{-0.35cm}

\[
g = \max_{i=1,\dots,N} \phi(x_i),
\]

\vspace{-0.1cm}

\noindent
where \( \phi \) is a shared MLP and \( g \) is a global feature vector summarizing the shape. This global descriptor can be directly used for classification. For segmentation, the global feature is concatenated with local point features before being passed to another shared MLP, enabling point-wise predictions with contextual awareness.
PointNet also includes a learnable alignment module, the T-Net, which estimates an affine transformation to align the input points or features, regularized by \( \| I - T T^\top \|_F^2 \) to keep the transformation close to orthogonal and enhance robustness.

\vspace{0.2cm}

PointNet++ builds on PointNet by introducing a hierarchical structure that captures local geometric features at multiple scales. It organizes the input point cloud using a series of \textit{Set Abstraction (SA)} layers, each composed of a sampling step using Farthest Point Sampling (FPS) to select representative points, followed by a grouping step that finds local neighborhoods around each sampled point. Each neighborhood is then processed with a PointNet-like architecture to extract local features.
This process builds progressively more abstract feature representations while preserving spatial locality. To enable dense predictions for tasks like segmentation, PointNet++ uses \textit{Feature Propagation} layers that interpolate features back to the original resolution, leveraging both local and global context through skip connections.
Thanks to its hierarchical design and ability to model non-uniform point densities, PointNet++ substantially outperforms PointNet in complex 3D tasks.
For the data, we used the meshes provided by the dataset and processed them as 3D objects to extract the surface features using TRELLIS \citep{xiang2024structured}. We encoded all the aneurysms and vessels from both the classification and segmentation datasets, resulting in 331 aneurysms and 1694 vessels, represented as point clouds with positions and 1024-dimensional features. Those point clouds were sampled to obtain a fixed number of points per object, which is necessary for the neural networks we used. We did three different samplings: 512, 1024, and 2048 points per object, following the approach in Intra3D \citep{yang2020intra}. 

\subsection{Classification}

To comprehensively evaluate the utility of the extracted features, we designed a series of classification experiments on the Intra3D dataset \citep{yang2020intra}. We employed a spectrum of models, ranging from established 3D deep learning architectures to simpler classical methods, allowing us to probe the features from multiple perspectives. The selected models include PointNet \citep{pointnet}, PointNet++ \citep{pointnetpp}, a standard Multi-Layer Perceptron (MLP), and finally, MLP and logistic regression models trained directly on Principal Component Analysis (PCA) projections of the feature space.
For the primary deep learning models, PointNet and PointNet++, we conducted comparative experiments. First, to establish a baseline, the models were trained using only the geometric priors provided in the dataset: 3D point coordinates and their corresponding surface normals. Subsequently, we performed identical training runs where the surface normals were replaced with our extracted TRELLIS features \citep{xiang2024structured}, enabling a direct assessment of the features' contribution.
Our implementation of PointNet \citep{pointnet} utilized a modified architecture incorporating message passing for feature aggregation, while omitting the original T-net module for transformation invariance. This adjustment encourages the model to rely more on the extracted features and less on raw spatial coordinates, allowing it to focus on high-level geometric representations provided by TRELLIS features. The network consisted of five modified PointNet layers, each comprising two 64-neuron ReLU-activated layers, followed by a global max-pooling operation and a final linear classifier. In contrast, for PointNet++ \citep{pointnetpp}, we implemented the original architecture precisely as described by the authors, maintaining its hierarchical feature learning structure. The detailed hyperparameters for both architectures are provided in the appendix.
To isolate and assess the intrinsic discriminative power of the TRELLIS features, we trained a simple MLP exclusively on the feature embeddings, deliberately excluding all spatial coordinate information. This ablation study helps to quantify the richness of the feature representation itself. The architecture consisted of two main blocks: a point-wise MLP shared across all points to process individual feature vectors, and a global MLP to aggregate these representations for a final classification.
Furthermore, to analyze the global structure of the feature space, we applied PCA to the aggregated feature vectors of each 3D object. We then trained both an MLP and a logistic regression model on low-dimensional point cloud representations derived from the principal components, using metrics such as the mean, standard deviation, minimum, and maximum of the components. This analysis serves to demonstrate how effectively the features encapsulate the object's class identity.
All deep learning models were trained using the AdamW \cite{loshchilov2019decoupledweightdecayregularization} optimizer with a learning rate of 0.001 and a cosine weight decay schedule. Models incorporating TRELLIS features converged rapidly, requiring only 20 epochs, whereas baseline models were trained for 100 epochs to ensure convergence. A batch size of 16 was used throughout.
\subsection{Segmentation}

For the task of per-point semantic segmentation, we adapted the PointNet \citep{pointnet} and PointNet++ \citep{pointnetpp} architectures. As with the classification task, our primary goal was to compare the model performance when using baseline geometric features (point coordinates and normals) against the performance achieved by augmenting the input with TRELLIS features.
The PointNet \citep{pointnet} segmentation model was derived from our modified classification architecture. The global pooling and final classification layers were replaced with upsampling and per-point prediction heads to produce a segmentation mask for the entire input cloud. The architecture consists of five PointNet layers with 32-neuron ReLU-activated sub-layers, followed by a final linear layer for two-class output. For PointNet++ \citep{pointnetpp}, we employed the original segmentation architecture, which leverages its hierarchical structure with feature propagation layers to combine local and global information for dense point-wise prediction effectively.
The training protocol for segmentation mirrored that of classification. We used the AdamW \cite{loshchilov2019decoupledweightdecayregularization} optimizer with a learning rate of 0.001 and a cosine weight decay schedule, with a batch size of 8. Models trained with TRELLIS features required 100 epochs for convergence, while the baseline models without these features were trained for 200 epochs.

\section{Graph Neural Networks (GNNs)} \label{GNN}

\subsection{Algorithm}

Graph Neural Networks (GNNs) are a class of neural networks designed to process graph-structured data that are particularly well-suited for tasks where the data can be represented with nodes and edges, such as 3D meshes.
The main idea behind GNNs is the message-passing mechanism that allows nodes in a graph to exchange information with their neighbors. On each iteration or layer, nodes compute messages based on their features and the features of their neighbors, and then, these messages are aggregated to update the node's representation. This process can be repeated for multiple layers, allowing nodes to gather information from increasingly distant neighbors.
This concept has been adapted for 3D aneurysm modeling to simulate hemodynamics in \citep{graphphysics} with an encode-process-decode framework that facilitates the application of GNNs on mesh structures, enabling the simulation of blood flow in aneurysms. Simulation performance is further enhanced by attention mechanisms \citep{VaswaniSPUJGKP17}, which focus computation on the most relevant mesh regions, thereby increasing both accuracy and efficiency.
For the simulation, \citep{graphphysics} represents the mesh by an undirected graph \( G = (V, E) \) with nodes \( V = \{x_i\}_{i=1}^N \), each \( x_i \in \mathbb{R}^p \), so the input is the matrix \( X \in \mathbb{R}^{N \times p} \). The model follows an encode-process-decode architecture: the encoder maps \( X \) into a latent space \( Z_0 = \text{MLP}(X) \in \mathbb{R}^{N \times d} \) using two linear layers. The processor applies \( L \) transformer blocks, where each block uses masked multi-head self-attention with the adjacency matrix \( A \in \{0,1\}^{N \times N} \) as a mask, computed as
\[
\text{Attention}(Z) = \text{softmax}\left(\frac{QK^\top \odot A}{\sqrt{d}}\right)V,
\]
where \( Q, K, V \) are learned linear projections of \( Z \). A Gated MLP with Gaussian Error Linear Unit (GeLU) \citep{hendrycks2023gaussianerrorlinearunits} non-linearity follows, defined as
\[
Z = W_f \big( \text{GeLU}(W_l Z + b_l) \odot (W_r Z + b_r) \big) + b_f,
\]
With residual connections and RMS (Root Mean Square)  normalization \citep{zhang2019rootmeansquarelayer} after each sub-layer. Finally, the decoder maps \( Z_L \) back to the output space via two linear layers.
To enhance the receptive field and information flow, the adjacency matrix is augmented by dilated adjacency with \( k \)-hop neighbors, random edges added dynamically, and global attention connecting important nodes to all others.

\subsection{Graph Neural Networks for Blood Flow Simulation}

For blood flow simulation, we used the AnXplore dataset \citep{anxplore} and the GNN architecture proposed in \citep{graphphysics}, without modifying the core architecture. The primary modification was the inclusion of surface features from TRELLIS \citep{xiang2024structured}, combined with the original mesh features for five runs, and five others without those features, repeating the process for different model sizes.
For the TRELLIS features \citep{xiang2024structured}, we extracted the surface mesh from the first time step of each simulation, as the external aneurysm geometry does not change during the simulation.

\section{Results} \label{RESULTS}

In this section, we present the results of our experiments using different models for classification, segmentation, and blood flow simulation on the Intra3D dataset \citep{yang2020intra} and the AnXplore dataset \citep{anxplore}. 

\subsection{Classification Results}

The classification results are evaluated by training 5 times the models over 5 different seeds, and the results are the mean and standard deviation over the 5 runs. The segmentation results are evaluated using 5-fold cross-validation, and the results are the mean and standard deviation over the 5 folds.
The results of the classification experiments on the Intra3D dataset \citep{yang2020intra} are shown in Table \ref{tab:classification_results}. We compare the performance of PointNet \citep{pointnet}, PointNet++ \citep{pointnetpp}, MLP, and PCA-based methods, with the best results from the Intra3D study \citep{yang2020intra} using PointNet++ \citep{pointnet} and PointCNN \citep{pointcnn}, and with the results from 3DMedPT \citep{yu20213dmedicalpointtransformer}. The results show that using TRELLIS features significantly improves the performance of both PointNet and PointNet++ compared to using only point coordinates and normals. Globally, the accuracy on vessels increases by approximately 8\%, the accuracy on aneurysms by approximately 95\% and the F1-score by between 10\% and 20\%, depending on the model. We also find that the MLP model performs well when trained on TRELLIS features, achieving high accuracy.
Our best results are obtained with PointNet++ using TRELLIS features on 512 samples, reaching a vessel accuracy of 99.88\%, an aneurysm accuracy of 100\%, and an F1 score of 0.999. These results show that with TRELLIS features, we can outperform state-of-the-art models on the Intra3D dataset \citep{yang2020intra} for all the metrics and using different models, even with classic neural networks and without relying on point cloud processing architectures.

\begin{table}[h!]
  \centering
  \resizebox{0.99\textwidth}{!}{
  \begin{tabular}{llccccccc}
  \toprule
  \textbf{Model} & \textbf{Features} & \textbf{Input} 
  & \multicolumn{2}{c}{\textbf{V. (\%)}} 
  & \multicolumn{2}{c}{\textbf{A. (\%)}} 
  & \multicolumn{2}{c}{\textbf{F1}} \\
  \cmidrule(lr){4-5} \cmidrule(lr){6-7} \cmidrule(lr){8-9}
  & & & mean & std & mean & std & mean & std \\
  \midrule
  \multirow{3}{*}{MLP} & \multirow{3}{*}{with}
    & 512 & 88.80 & 24.55 & 97.84 & 1.55 & 0.9127 & 0.1827 \\
    &  & 1024 & \textbf{99.94} & \textbf{0.13} & 98.52 & 2.05 & 0.9970 & 0.0033 \\
    &  & 2048 & 99.82 & 0.16 & 97.87 & 1.82 & 0.9950 & 0.0039 \\
    \midrule
    \multirow{3}{*}{PointNet \citep{pointnet}} & \multirow{3}{*}{without}
    & 512 & 91.79 & 3.26 & 52.09 & 8.59 & 0.8209 & 0.0280 \\
    &  & 1024 & 92.48 & 2.06 & 52.33 & 4.52 & 0.8274 & 0.0059 \\
    &  & 2048 & 91.70 & 1.90 & 45.17 & 5.93 & 0.8019 & 0.0165 \\
    \midrule
    \multirow{3}{*}{PointNet \citep{pointnet}} & \multirow{3}{*}{with}
    & 512 & \textbf{99.94} & 1.29 & 99.40 & 0.82 & 0.9985 & \textbf{0.0014} \\
    &  & 1024 & 99.82 & 0.27 & 98.15 & 0.78 & 0.9956 & 0.0021 \\
    &  & 2048 & 99.83 & 0.38 & 99.14 & 1.29 & 0.9971 & 0.0032 \\
    \midrule
    \multirow{3}{*}{PointNet++ \citep{pointnetpp}} & \multirow{3}{*}{without}
    & 512 & 90.30 & 2.04 & 52.59 & 2.61 & 0.8130 & 0.0124 \\
    &  & 1024 & 89.69 & 2.23 & 53.01 & 2.36 & 0.8106 & 0.0123 \\
    &  & 2048 & 90.57 & 2.28 & 51.29 & 4.49 & 0.8105 & 0.0155 \\
    \midrule
    \multirow{3}{*}{PointNet++ \citep{pointnetpp}} & \multirow{3}{*}{with}
    & 512 & 99.88 & 0.16 & \textbf{100} & \textbf{0.00} & \textbf{0.9990} & \textbf{0.0014} \\
    &  & 1024 & 99.76 & 0.32 & 98.79 & 1.28 & 0.9961 & 0.0028 \\
    &  & 2048 & 99.46 & 0.25 & 97.96 & 2.16 & 0.9921 & 0.0033 \\
    \midrule
  \multirow{3}{*}{\shortstack{PointNet++ \citep{pointnetpp} \\(from \citep{yang2020intra})}} & \multirow{3}{*}{without}
  & 512 & 98.52 & - & 86.69 & - & 0.8928 & -\\
  & & 1024 & 98.52 & - & 88.51 & - & 0.9029 & - \\
  & & 2048 & 98.76 & - & 87.31 & - & 0.9016 & - \\
  \midrule
  \multirow{3}{*}{\shortstack{PointCNN \citep{pointcnn} \\(from \citep{yang2020intra})}} & \multirow{3}{*}{without}
  & 512 & 98.38 & - & 78.25 & - & 0.8494 & -  \\
  & & 1024 & 98.79 & - & 81.28 & - & 0.8748 & - \\
  & & 2048 & 98.95 & - & 85.81 & - & 0.9044 & - \\
  \midrule
  \multirow{3}{*}{\shortstack{3DMedPT \citep{yu20213dmedicalpointtransformer}}} & \multirow{3}{*}{without}
  & 512 & 99.02 & - & 94.06 & - & 0.920 & -  \\
  & & 1024 & 99.24 & - & 93.26 & - & 0.936 & - \\
  & & 2048 & 99.07 & - & 93.49 & - & 0.931 & - \\
  \midrule
  \midrule
  \multirow{3}{*}{MLP on PCA} 
  & mean & & 94.21 & 2.79 & 75.20 & 9.27 & 0.9117 & 0.0066\\
  & mean and std& & 97.93 & 1.55 & 54.17 & 17.75 & 0.8962 & 0.0302\\
  & all metrics & & 96.87 & 1.37 & 80.98 & 3.93 & 0.9424 & 0.0131\\
  \midrule
  \multirow{3}{*}{\shortstack{Logistic Regression \\ on PCA}}    
  & mean & & 95.75 & 0.668 & 65.57 & 4.35 & 0.9054 & 0.0414\\
  & mean and std & & 95.98 & 0.82 & 68.58 & 3.37 & 0.9130 & 0.0059\\
  & all metrics & & 97.18 & 1.16 & 78.26 & 4.36 & 0.9398 & 0.0116 \\
  \bottomrule
  \end{tabular}}
  \caption{Full comparison of the different models for classification on the Intra3D dataset. Results are mean and standard deviation (std) on vessels segment accuracy (V.), aneurysms segment accuracy (A.), and F1-score.}
  \label{tab:classification_results}
\end{table}

\subsection{Segmentation Results}

The segmentation results on the Intra3D dataset \citep{yang2020intra} are shown in Table \ref{tab:segmentation_results}. We compare the performance of PointNet \citep{pointnet} and PointNet++ \citep{pointnetpp} models, both with and without TRELLIS features, with the best results from the Intra3D study \citep{yang2020intra} using PointNet++ \citep{pointnetpp} and SO-net \citep{sonet}, and with the results from 3DMedPT \citep{yu20213dmedicalpointtransformer}. The results indicate that both models achieve high accuracy in segmenting aneurysms and healthy vessels. Furthermore, the addition of TRELLIS features consistently leads to notable improvements in segmentation performance when compared to benchmark results.
Also, all the algorithms using TRELLIS features outperform the best results from the Intra3D study \citep{yang2020intra} and 3DMedPT \citep{yu20213dmedicalpointtransformer} for all metrics, including IoU (Intersection of Union) and DSC (Dice Similarity Coefficient).
The results show that TRELLIS features significantly enhance the segmentation performance of both PointNet and PointNet++. 

\begin{table}[h!]
  \centering
  \resizebox{0.99\textwidth}{!}{
  \begin{tabular}{llccccccccc}
  \toprule
  \textbf{Model} & \textbf{Features} & \textbf{Input} 
  & \multicolumn{2}{c}{\textbf{IoU V. (\%)}} 
  & \multicolumn{2}{c}{\textbf{IoU A. (\%)}} 
  & \multicolumn{2}{c}{\textbf{DSC V. (\%)}}
  & \multicolumn{2}{c}{\textbf{DSC A. (\%)}} \\
  \cmidrule(lr){4-5} \cmidrule(lr){6-7} \cmidrule(lr){8-9} \cmidrule(lr){10-11}
  & & & mean & std & mean & std & mean & std & mean & std \\
  \midrule
  \multirow{3}{*}{PointNet \citep{pointnet}} & \multirow{3}{*}{without}
  & 512 & 88.08 & 1.74 & 66.38 & 3.84 & 93.65 & 0.99 & 79.75 & 2.54 \\
  &  & 1024 & 85.81 & 1.13 & 60.17 & 3.31 & 92.18 & 0.66 & 75.09 & 2.64 \\
  &  & 2048 & 81.16 & 2.04 & 50.95 & 7.97 & 89.59 & 1.26 & 67.19 & 7.45 \\ 
  \midrule
  \multirow{3}{*}{PointNet \citep{pointnet}} & \multirow{3}{*}{with}
  & 512 & 95.84 & 0.37 & 86.53 & 2.40 & 97.87 & 0.16 & 92.77 & 1.29 \\
  &  & 1024 & 96.53 & 0.41 & 88.59 & 1.86 & 98.23 & 0.22 & 93.94 & 1.05 \\
  &  & 2048 & \textbf{96.57} & 0.28 & \textbf{88.67} & \textbf{1.82} & \textbf{98.26} & 0.15 & \textbf{93.99} & \textbf{1.03} \\
  \midrule
  \multirow{3}{*}{PointNet++ \citep{pointnetpp}} & \multirow{3}{*}{without}
  & 512 & 87.87 & 2.80 & 64.75 & 5.50 & 93.52 & 1.62 & 78.50 & 4.05 \\
  &  & 1024 & 88.52 & 1.08 & 66.68 & 5.67 & 93.91 & 0.60 & 79.90 & 4.17\\
  &  & 2048 & 89.33 & 1.21 & 67.98 & 3.69 & 94.30 & 0.68 & 80.89 & 2.60\\
  \midrule
  \multirow{3}{*}{PointNet++ \citep{pointnetpp}} & \multirow{3}{*}{with}
  & 512 & 96.13 & 0.44 & 87.19 & 2.62 & 98.03 & 0.23 & 93.14 & 1.52 \\
  &  & 1024 & 96.38 & \textbf{0.22} & 88.09 & 2.17 & 98.15 & \textbf{0.11} & 93.66 & 1.23 \\
  &  & 2048 & 96.46 & 0.40 & 88.31 & 2.51 & 98.20 & 0.21 & 93.78 & 1.43 \\
  \midrule
  \multirow{3}{*}{\shortstack{PointNet++ \citep{pointnetpp} \\ (from \citep{yang2020intra})}} & \multirow{3}{*}{without}
  & 512 & 93.34 & 1.28 & 76.22 & 4.85 & 96.48 & 0.73 & 83.92 & 4.34 \\
  & & 1024 & 93.35 & 1.15 & 76.38 & 4.36 &96.47 & 0.65 & 84.62 & 3.82 \\ 
  & & 2048 & 93.24 & 1.18 & 76.21 & 4.34 & 96.40 & 0.67 & 84.64 & 3.77 \\
  \midrule
  \multirow{3}{*}{\shortstack{SO-net \citep{sonet} \\ (from \citep{yang2020intra})}} & \multirow{3}{*}{without}
  & 512 & 94.22 & 1.07 & 80.14 & 3.28 & 96.95 & 0.59 & 87.90 & 2.43 \\
  &  & 1024 & 94.42 & 1.04 & 80.99 & 3.21 & 97.06 & 0.58 & 88.41 & 2.43 \\
  &  & 2048 & 94.46 & 1.00 & 81.40 & 3.09 & 97.09 & 0.55 & 88.76 & 2.24 \\
  \midrule
  \multirow{3}{*}{\shortstack{3DMedPT \citep{yu20213dmedicalpointtransformer}}} & \multirow{3}{*}{without}
  & 512 & 94.82 & - & 81.80 & - & 97.29 & - & 89.25 & - \\
  & & 1024 & 94.76 & - & 82.39 & - & 97.25 & - & 89.71 & - \\
  & & 2048 & 93.52 & - & 80.13 & - & 96.59 & - & 88.69 & - \\
  \bottomrule
  \end{tabular}}
  \caption{Full comparison of the different models for segmentation on the Intra3D dataset. Results are mean and standard deviation (std) on vessels segment accuracy (V.), aneurysms segment accuracy (A.), and F1-score.}
  \label{tab:segmentation_results}
\end{table}

\subsection{Blood Flow Simulation Results}

The results of the GNN + Transformers model \citep{graphphysics} on the AnXplore dataset \citep{anxplore} are shown in Table \ref{tab:gnn_results}. Our model was trained using the features extracted from TRELLIS \citep{xiang2024structured} and the mesh structure of the aneurysms, and we find that the model achieves a lower RMSE (Root Mean Square Error) across all time steps of the blood flow simulation with the addition of TRELLIS features compared to using only the original features (point coordinates and normals). The error is reduced by approximately 15\%, demonstrating the effectiveness of these features for simulating blood flow.

\begin{table}[h!]
  \centering
  \setlength{\tabcolsep}{1.06em} 
  \renewcommand{\arraystretch}{1.12} 
  \begin{tabular}{lcc}
  \toprule
  \multirow{2}{*}{\textbf{Model}} & \multicolumn{2}{c}{\textbf{All-Rollout RMSE}} \\
  \cmidrule(lr){2-3}
  & \textbf{Mean} & \textbf{Std} \\
  \midrule
 S/1 & 7.57 & 1.103 \\
  \midrule
 S/1 + feats & 6.09 & 0.637 \\
  \midrule
 L/1 & 4.03 & 0.330 \\
  \midrule
 L/1 + feats & \textbf{3.55} & 0.170 \\
  \bottomrule
  \end{tabular}
  \caption{Results of the GNN + Transformers model for \textbf{simulation} on the AnXplore dataset. The All-Rollout RMSE is computed over all time steps of the blood flow simulation.}
  \label{tab:gnn_results}
\end{table}

\subsection{Discussion on the Computational Cost}

However, this approach impacts the training time for the models. Processing the size 1024 vectors used as features for each point significantly increases the training time, as the model must process a larger number of features compared to using only point coordinates and normals. It multiplies the training time by a factor of two for the segmentation and by five for the classification tasks.
Furthermore, the encoding time is significant: encoding 400 objects with TRELLIS on an A100 GPU takes about 12 hours, which averages to five minutes per object. While this is a significant amount of time, it is a one-time process. After encoding, we can train different models on the resulting dataset. In comparison, running the entire pipeline of encoding and training on the full dataset would increase the total time by a factor of 30 compared to training point cloud processing models such as PointNet or PointNet++ alone. The major time-consuming step is the rendering process, taking 200 views per object with Blender, representing around 80\% of the total time. Reducing this number to 100 or even 50 views could significantly decrease the encoding time, making it drop to 40 to 60\% of the current time, which would be more manageable. Still, an in-depth study is needed to determine how important the number of views is for the overall performance of the models. A balance must be found between the number of views and the performance of the models.

\section{Conclusion} \label{CONCLUSION}

In this work, we have demonstrated the remarkable efficacy of transferring latent geometric knowledge from a general-purpose 3D generative model to specialized medical imaging tasks. By leveraging features extracted from the TRELLIS encoder \citep{xiang2024structured}, we have shown significant performance enhancements in the classification and segmentation of intracranial aneurysms from the Intra3D dataset \citep{yang2020intra}, as well as in the simulation of intra-aneurysmal hemodynamics on the AnXplore dataset \citep{anxplore}. Our results establish a new state-of-the-art on the Intra3D benchmark, with augmented PointNet \citep{pointnet} and PointNet++ \citep{pointnetpp} architectures substantially outperforming models trained solely on point coordinates and normals.
A key finding of this study is the sheer potency of these pre-trained features. A simple multi-layer perceptron, devoid of any specialized 3D processing architecture, achieved competitive and even superior results when trained on TRELLIS embeddings alone. This suggests that the encoder, despite its training on non-medical data, distills a rich and highly informative representation of 3D shape that is directly relevant to cerebrovascular pathology. Our feature analysis corroborates this, revealing that the latent space effectively organizes aneurysms by salient geometric characteristics. This property holds significant potential for future clinical applications, such as identifying morphological markers associated with rupture risk.
The success of this cross-domain strategy presents a promising new direction for machine learning in medical imaging. The vast and diverse datasets used to train foundational generative models can serve as a powerful source of prior knowledge, helping to overcome the data scarcity that often plagues medical research. The ability of the TRELLIS encoder to capture complex 3D structures opens new avenues for enhancing a wide array of computational models in medicine.
Looking forward, several promising research directions emerge from this work. A natural extension is to investigate the generalizability of this approach to other 3D medical datasets, such as different modalities within the MedMNIST v2 collection \citep{Yang_2023}, and to explore its synergy with other network architectures like SO-net \citep{sonet} or PointCNN \citep{pointcnn}. Furthermore, although our results are compelling, the performance could be further elevated through fine-tuning. A targeted fine-tuning of the TRELLIS encoder on a large-scale medical dataset, such as MedMNIST v2, could adapt its feature space to be even more discriminative for biomedical structures. Finally, practical implementation will require optimizing the feature extraction process. A systematic study on the trade-off between the number of rendering views used for encoding and the final model performance could lead to a more computationally efficient pipeline without compromising diagnostic accuracy.
In conclusion, this study validates the use of features from large-scale generative models as a potent tool for enhancing 3D medical data analysis. By bridging the gap between generalist 3D representation learning and specialized medical diagnostics, our work paves the way for more accurate, robust, and data-efficient models in the fight against life-threatening conditions, such as intracranial aneurysms.

\section{Acknowledgements}

\noindent The authors acknowledge the financial support from ERC grant no 2021-CoG-101045042, CURE. Views and opinions expressed are however those of the author(s) only and do not necessarily reflect those of the European Union or the European Research Council. Neither the European Union nor the granting authority can be held responsible for them.

\bibliographystyle{unsrtnat}
\bibliography{biblio}

\begin{thebibliography}{28}
\providecommand{\natexlab}[1]{#1}
\providecommand{\url}[1]{\texttt{#1}}
\expandafter\ifx\csname urlstyle\endcsname\relax
  \providecommand{\doi}[1]{doi: #1}\else
  \providecommand{\doi}{doi: \begingroup \urlstyle{rm}\Url}\fi

\bibitem[Yang et~al.(2020)Yang, Xia, Kin, and Igarashi]{yang2020intra}
Xi~Yang, Ding Xia, Taichi Kin, and Takeo Igarashi.
\newblock Intra: 3d intracranial aneurysm dataset for deep learning.
\newblock In \emph{The IEEE Conference on Computer Vision and Pattern
  Recognition (CVPR)}, 2020.

\bibitem[Yang et~al.(2023)Yang, Shi, Wei, Liu, Zhao, Ke, Pfister, and
  Ni]{Yang_2023}
Jiancheng Yang, Rui Shi, Donglai Wei, Zequan Liu, Lin Zhao, Bilian Ke,
  Hanspeter Pfister, and Bingbing Ni.
\newblock Medmnist v2 - a large-scale lightweight benchmark for 2d and 3d
  biomedical image classification.
\newblock \emph{Scientific Data}, 10\penalty0 (1), January 2023.
\newblock ISSN 2052-4463.
\newblock \doi{10.1038/s41597-022-01721-8}.
\newblock URL \url{http://dx.doi.org/10.1038/s41597-022-01721-8}.

\bibitem[Yu et~al.(2021)Yu, Zhang, Wang, Zhang, Song, Xiang, Liu, and
  Cai]{yu20213dmedicalpointtransformer}
Jianhui Yu, Chaoyi Zhang, Heng Wang, Dingxin Zhang, Yang Song, Tiange Xiang,
  Dongnan Liu, and Weidong Cai.
\newblock 3d medical point transformer: Introducing convolution to attention
  networks for medical point cloud analysis, 2021.
\newblock URL \url{https://arxiv.org/abs/2112.04863}.

\bibitem[Liu et~al.(2023)Liu, Li, Liu, Chen, and Yuan]{10093984}
Yifan Liu, Wuyang Li, Jie Liu, Hui Chen, and Yixuan Yuan.
\newblock Grab-net: Graph-based boundary-aware network for medical point cloud
  segmentation.
\newblock \emph{IEEE Transactions on Medical Imaging}, 42\penalty0
  (9):\penalty0 2776--2786, 2023.
\newblock \doi{10.1109/TMI.2023.3265000}.

\bibitem[Arzani et~al.(2021)Arzani, Wang, and D’Souza]{Arzani_2021}
Amirhossein Arzani, Jian-Xun Wang, and Roshan~M. D’Souza.
\newblock Uncovering near-wall blood flow from sparse data with
  physics-informed neural networks.
\newblock \emph{Physics of Fluids}, 33\penalty0 (7), July 2021.
\newblock ISSN 1089-7666.
\newblock \doi{10.1063/5.0055600}.
\newblock URL \url{http://dx.doi.org/10.1063/5.0055600}.

\bibitem[Suk et~al.(2024)Suk, de~Haan, Lippe, Brune, and Wolterink]{Suk_2024}
Julian Suk, Pim de~Haan, Phillip Lippe, Christoph Brune, and Jelmer~M.
  Wolterink.
\newblock Mesh neural networks for se(3)-equivariant hemodynamics estimation on
  the artery wall.
\newblock \emph{Computers in Biology and Medicine}, 173:\penalty0 108328, May
  2024.
\newblock ISSN 0010-4825.
\newblock \doi{10.1016/j.compbiomed.2024.108328}.
\newblock URL \url{http://dx.doi.org/10.1016/j.compbiomed.2024.108328}.

\bibitem[Garnier et~al.(2025)Garnier, Lannelongue, Viquerat, and
  Hachem]{graphphysics}
Paul Garnier, Vincent Lannelongue, Jonathan Viquerat, and Elie Hachem.
\newblock Training transformers for mesh-based simulations, 2025.
\newblock URL \url{https://arxiv.org/abs/2508.18051}.

\bibitem[A et~al.(2024)A, P, U, Y, J, and E]{anxplore}
Goetz A, Jeken-Rico P, Pelissier U, Chau Y, Sédat J, and Hachem E.
\newblock Anxplore: a comprehensive fluid-structure interaction study of 101
  intracranial aneurysms.
\newblock 2024.

\bibitem[Pfaff et~al.(2021)Pfaff, Fortunato, Sanchez-Gonzalez, and
  Battaglia]{pfaff2021learningmeshbasedsimulationgraph}
Tobias Pfaff, Meire Fortunato, Alvaro Sanchez-Gonzalez, and Peter~W. Battaglia.
\newblock Learning mesh-based simulation with graph networks.
\newblock 2021.
\newblock URL \url{https://arxiv.org/abs/2010.03409}.

\bibitem[Vaswani et~al.(2017)Vaswani, Shazeer, Parmar, Uszkoreit, Jones, Gomez,
  Kaiser, and Polosukhin]{VaswaniSPUJGKP17}
Ashish Vaswani, Noam Shazeer, Niki Parmar, Jakob Uszkoreit, Llion Jones,
  Aidan~N. Gomez, Lukasz Kaiser, and Illia Polosukhin.
\newblock Attention is all you need.
\newblock \emph{CoRR}, abs/1706.03762, 2017.
\newblock URL \url{http://arxiv.org/abs/1706.03762}.

\bibitem[Xiang et~al.(2024)Xiang, Lv, Xu, Deng, Wang, Zhang, Chen, Tong, and
  Yang]{xiang2024structured}
Jianfeng Xiang, Zelong Lv, Sicheng Xu, Yu~Deng, Ruicheng Wang, Bowen Zhang,
  Dong Chen, Xin Tong, and Jiaolong Yang.
\newblock Structured 3d latents for scalable and versatile 3d generation.
\newblock \emph{arXiv preprint arXiv:2412.01506}, 2024.

\bibitem[Ling et~al.(2025)Ling, Owalekar, Adesanya, Bıyık, and
  Seita]{ling2025impactintelligentmotionplanning}
Yiyang Ling, Karan Owalekar, Oluwatobiloba Adesanya, Erdem Bıyık, and Daniel
  Seita.
\newblock Impact: Intelligent motion planning with acceptable contact
  trajectories via vision-language models, 2025.
\newblock URL \url{https://arxiv.org/abs/2503.10110}.

\bibitem[Jiang et~al.(2025)Jiang, Hsu, Zhang, Yu, Wang, and
  Li]{jiang2025phystwinphysicsinformedreconstructionsimulation}
Hanxiao Jiang, Hao-Yu Hsu, Kaifeng Zhang, Hsin-Ni Yu, Shenlong Wang, and Yunzhu
  Li.
\newblock Phystwin: Physics-informed reconstruction and simulation of
  deformable objects from videos, 2025.
\newblock URL \url{https://arxiv.org/abs/2503.17973}.

\bibitem[Deitke et~al.(2023)Deitke, Liu, Wallingford, Ngo, Michel, Kusupati,
  Fan, Laforte, Voleti, Gadre, VanderBilt, Kembhavi, Vondrick, Gkioxari,
  Ehsani, Schmidt, and Farhadi]{deitke2023objaversexluniverse10m3d}
Matt Deitke, Ruoshi Liu, Matthew Wallingford, Huong Ngo, Oscar Michel, Aditya
  Kusupati, Alan Fan, Christian Laforte, Vikram Voleti, Samir~Yitzhak Gadre,
  Eli VanderBilt, Aniruddha Kembhavi, Carl Vondrick, Georgia Gkioxari, Kiana
  Ehsani, Ludwig Schmidt, and Ali Farhadi.
\newblock Objaverse-xl: A universe of 10m+ 3d objects, 2023.
\newblock URL \url{https://arxiv.org/abs/2307.05663}.

\bibitem[Collins et~al.(2022)Collins, Goel, Deng, Luthra, Xu, Gundogdu, Zhang,
  Vicente, Dideriksen, Arora, Guillaumin, and
  Malik]{collins2022abodatasetbenchmarksrealworld}
Jasmine Collins, Shubham Goel, Kenan Deng, Achleshwar Luthra, Leon Xu, Erhan
  Gundogdu, Xi~Zhang, Tomas F.~Yago Vicente, Thomas Dideriksen, Himanshu Arora,
  Matthieu Guillaumin, and Jitendra Malik.
\newblock Abo: Dataset and benchmarks for real-world 3d object understanding,
  2022.
\newblock URL \url{https://arxiv.org/abs/2110.06199}.

\bibitem[Fu et~al.(2020)Fu, Jia, Gao, Gong, Zhao, Maybank, and
  Tao]{fu20203dfuture3dfurnitureshape}
Huan Fu, Rongfei Jia, Lin Gao, Mingming Gong, Binqiang Zhao, Steve Maybank, and
  Dacheng Tao.
\newblock 3d-future: 3d furniture shape with texture, 2020.
\newblock URL \url{https://arxiv.org/abs/2009.09633}.

\bibitem[Khanna et~al.(2023)Khanna, Mao, Jiang, Haresh, Shacklett, Batra,
  Clegg, Undersander, Chang, and
  Savva]{khanna2023habitatsyntheticscenesdataset}
Mukul Khanna, Yongsen Mao, Hanxiao Jiang, Sanjay Haresh, Brennan Shacklett,
  Dhruv Batra, Alexander Clegg, Eric Undersander, Angel~X. Chang, and Manolis
  Savva.
\newblock Habitat synthetic scenes dataset (hssd-200): An analysis of 3d scene
  scale and realism tradeoffs for objectgoal navigation, 2023.
\newblock URL \url{https://arxiv.org/abs/2306.11290}.

\bibitem[Qi et~al.(2016)Qi, Su, Mo, and Guibas]{pointnet}
Charles~Ruizhongtai Qi, Hao Su, Kaichun Mo, and Leonidas~J. Guibas.
\newblock Pointnet: Deep learning on point sets for 3d classification and
  segmentation.
\newblock \emph{CoRR}, abs/1612.00593, 2016.
\newblock URL \url{http://arxiv.org/abs/1612.00593}.

\bibitem[Qi et~al.(2017)Qi, Yi, Su, and Guibas]{pointnetpp}
Charles~Ruizhongtai Qi, Li~Yi, Hao Su, and Leonidas~J. Guibas.
\newblock Pointnet++: Deep hierarchical feature learning on point sets in a
  metric space.
\newblock \emph{CoRR}, abs/1706.02413, 2017.
\newblock URL \url{http://arxiv.org/abs/1706.02413}.

\bibitem[Li et~al.(2018{\natexlab{a}})Li, Bu, Sun, and Chen]{pointcnn}
Yangyan Li, Rui Bu, Mingchao Sun, and Baoquan Chen.
\newblock Pointcnn.
\newblock \emph{CoRR}, abs/1801.07791, 2018{\natexlab{a}}.
\newblock URL \url{http://arxiv.org/abs/1801.07791}.

\bibitem[Li et~al.(2018{\natexlab{b}})Li, Chen, and Lee]{sonet}
Jiaxin Li, Ben~M. Chen, and Gim~Hee Lee.
\newblock So-net: Self-organizing network for point cloud analysis.
\newblock \emph{CoRR}, abs/1803.04249, 2018{\natexlab{b}}.
\newblock URL \url{http://arxiv.org/abs/1803.04249}.

\bibitem[Wu et~al.(2018)Wu, Qi, and Li]{pointconv}
Wenxuan Wu, Zhongang Qi, and Fuxin Li.
\newblock Pointconv: Deep convolutional networks on 3d point clouds.
\newblock \emph{CoRR}, abs/1811.07246, 2018.
\newblock URL \url{http://arxiv.org/abs/1811.07246}.

\bibitem[Oquab et~al.(2024)Oquab, Darcet, Moutakanni, Vo, Szafraniec, Khalidov,
  Fernandez, Haziza, Massa, El-Nouby, Assran, Ballas, Galuba, Howes, Huang, Li,
  Misra, Rabbat, Sharma, Synnaeve, Xu, Jegou, Mairal, Labatut, Joulin, and
  Bojanowski]{oquab2024dinov2learningrobustvisual}
Maxime Oquab, Timothée Darcet, Théo Moutakanni, Huy Vo, Marc Szafraniec,
  Vasil Khalidov, Pierre Fernandez, Daniel Haziza, Francisco Massa, Alaaeldin
  El-Nouby, Mahmoud Assran, Nicolas Ballas, Wojciech Galuba, Russell Howes,
  Po-Yao Huang, Shang-Wen Li, Ishan Misra, Michael Rabbat, Vasu Sharma, Gabriel
  Synnaeve, Hu~Xu, Hervé Jegou, Julien Mairal, Patrick Labatut, Armand Joulin,
  and Piotr Bojanowski.
\newblock Dinov2: Learning robust visual features without supervision, 2024.
\newblock URL \url{https://arxiv.org/abs/2304.07193}.

\bibitem[Dosovitskiy et~al.(2021)Dosovitskiy, Beyer, Kolesnikov, Weissenborn,
  Zhai, Unterthiner, Dehghani, Minderer, Heigold, Gelly, Uszkoreit, and
  Houlsby]{dosovitskiy2021imageworth16x16words}
Alexey Dosovitskiy, Lucas Beyer, Alexander Kolesnikov, Dirk Weissenborn,
  Xiaohua Zhai, Thomas Unterthiner, Mostafa Dehghani, Matthias Minderer, Georg
  Heigold, Sylvain Gelly, Jakob Uszkoreit, and Neil Houlsby.
\newblock An image is worth 16x16 words: Transformers for image recognition at
  scale, 2021.
\newblock URL \url{https://arxiv.org/abs/2010.11929}.

\bibitem[Wu et~al.(2015)Wu, Song, Khosla, Yu, Zhang, Tang, and Xiao]{7298801}
Zhirong Wu, Shuran Song, Aditya Khosla, Fisher Yu, Linguang Zhang, Xiaoou Tang,
  and Jianxiong Xiao.
\newblock 3d shapenets: A deep representation for volumetric shapes.
\newblock In \emph{2015 IEEE Conference on Computer Vision and Pattern
  Recognition (CVPR)}, pages 1912--1920, 2015.
\newblock \doi{10.1109/CVPR.2015.7298801}.

\bibitem[Loshchilov and
  Hutter(2019)]{loshchilov2019decoupledweightdecayregularization}
Ilya Loshchilov and Frank Hutter.
\newblock Decoupled weight decay regularization, 2019.
\newblock URL \url{https://arxiv.org/abs/1711.05101}.

\bibitem[Hendrycks and Gimpel(2023)]{hendrycks2023gaussianerrorlinearunits}
Dan Hendrycks and Kevin Gimpel.
\newblock Gaussian error linear units (gelus), 2023.
\newblock URL \url{https://arxiv.org/abs/1606.08415}.

\bibitem[Zhang and Sennrich(2019)]{zhang2019rootmeansquarelayer}
Biao Zhang and Rico Sennrich.
\newblock Root mean square layer normalization, 2019.
\newblock URL \url{https://arxiv.org/abs/1910.07467}.

\end{thebibliography}

\appendix

\clearpage
\section{Supplementary figures on the analysis of the TRELLIS features}
\label{annexeA}

\vspace{-0.3cm}

\begin{figure}[h!]
  \centering
  \includegraphics[width=0.32\textwidth]{pca_mean.png}
  \includegraphics[width=0.32\textwidth]{pca_std.png}\\
  \includegraphics[width=0.32\textwidth]{pca_min.png}
  \includegraphics[width=0.32\textwidth]{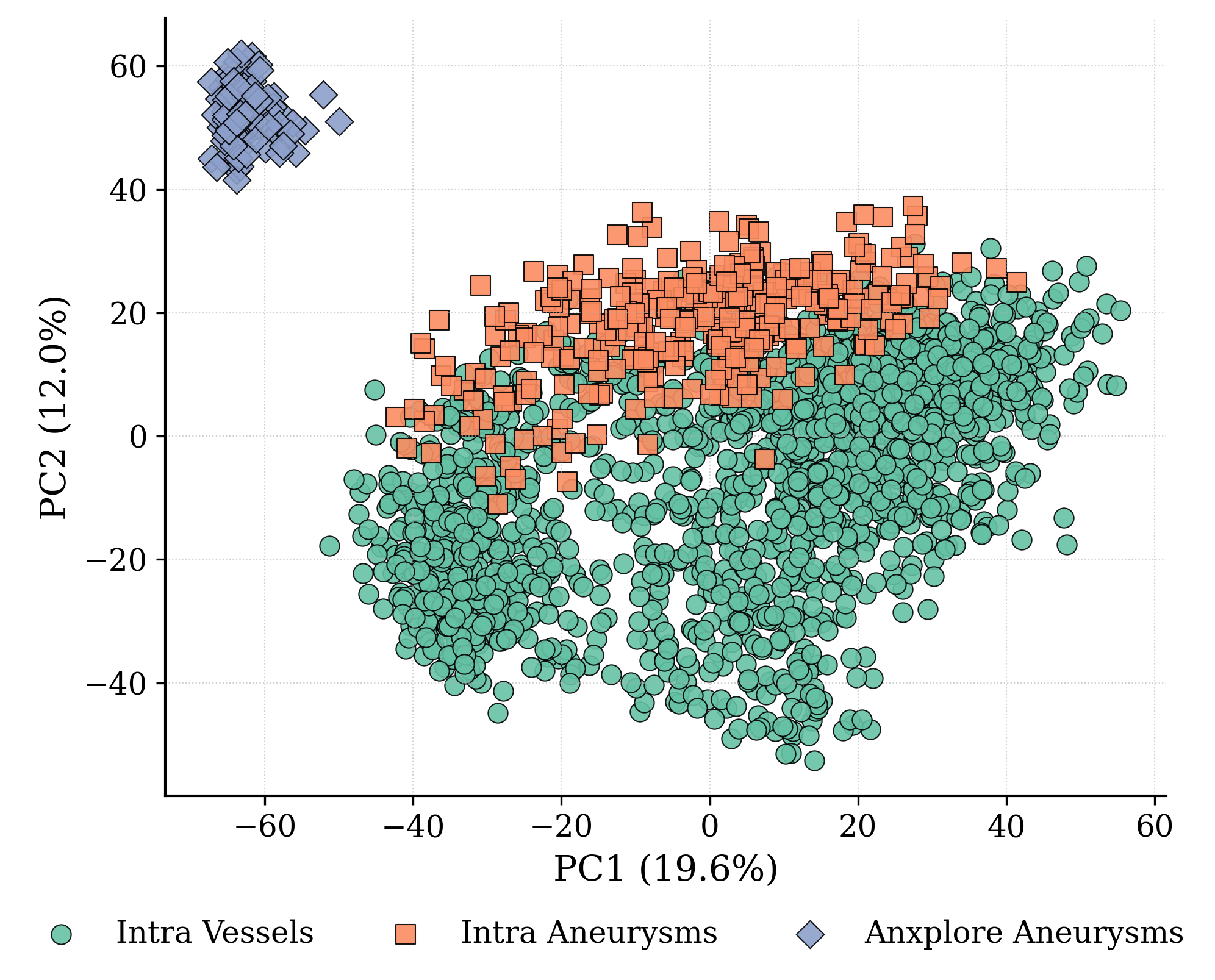}
  \caption{Results of PCA on the Intra3D dataset combined with the AnXplore dataset. Figures show the mean, standard deviation, minimum, and maximum of the PCA components.}
  \label{fig:pca_supplementary_all}
\end{figure}

\vspace{-0.55cm}

\begin{figure}[h!]
    \centering
    \includegraphics[width=0.32\textwidth]{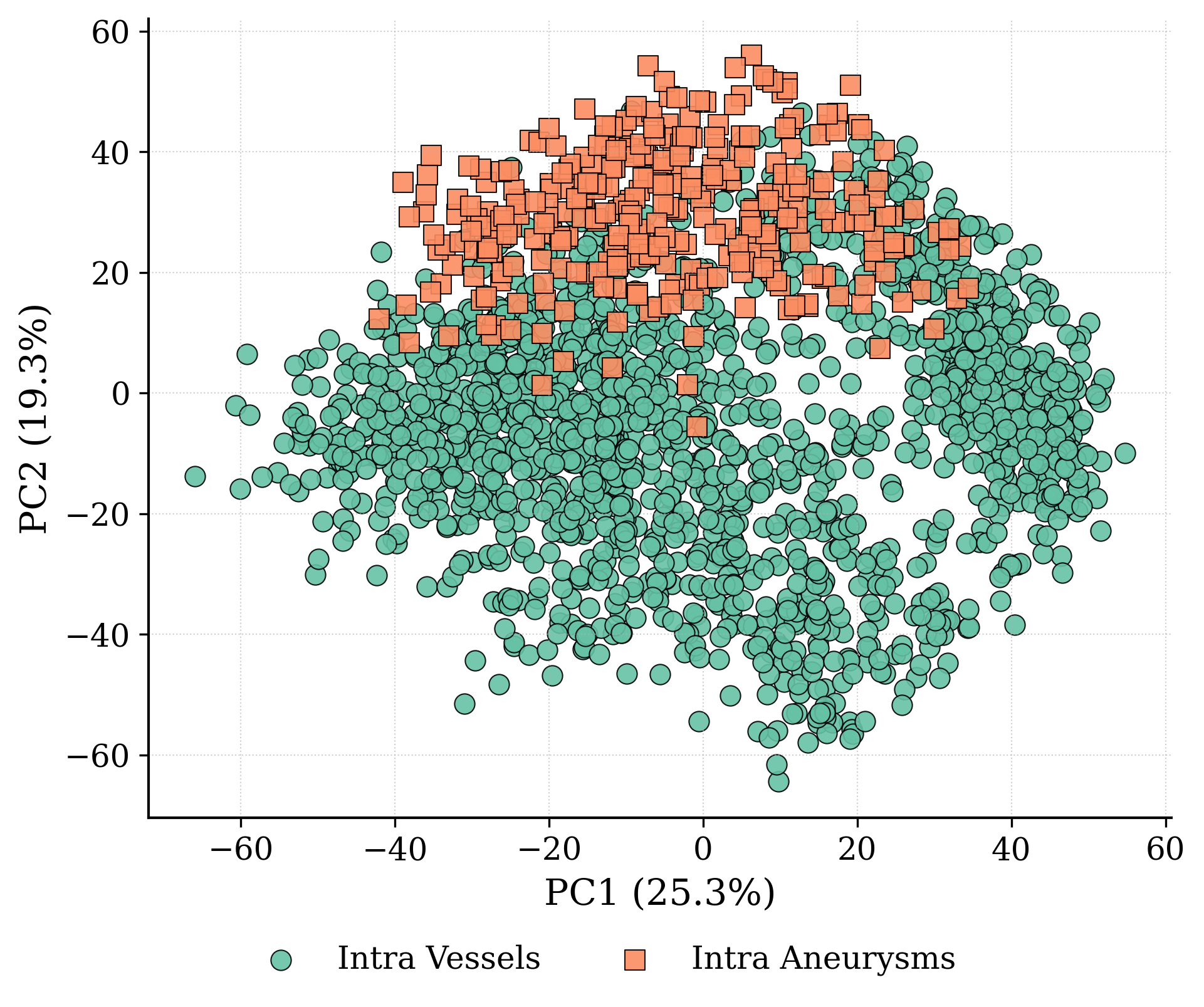}
    \includegraphics[width=0.32\textwidth]{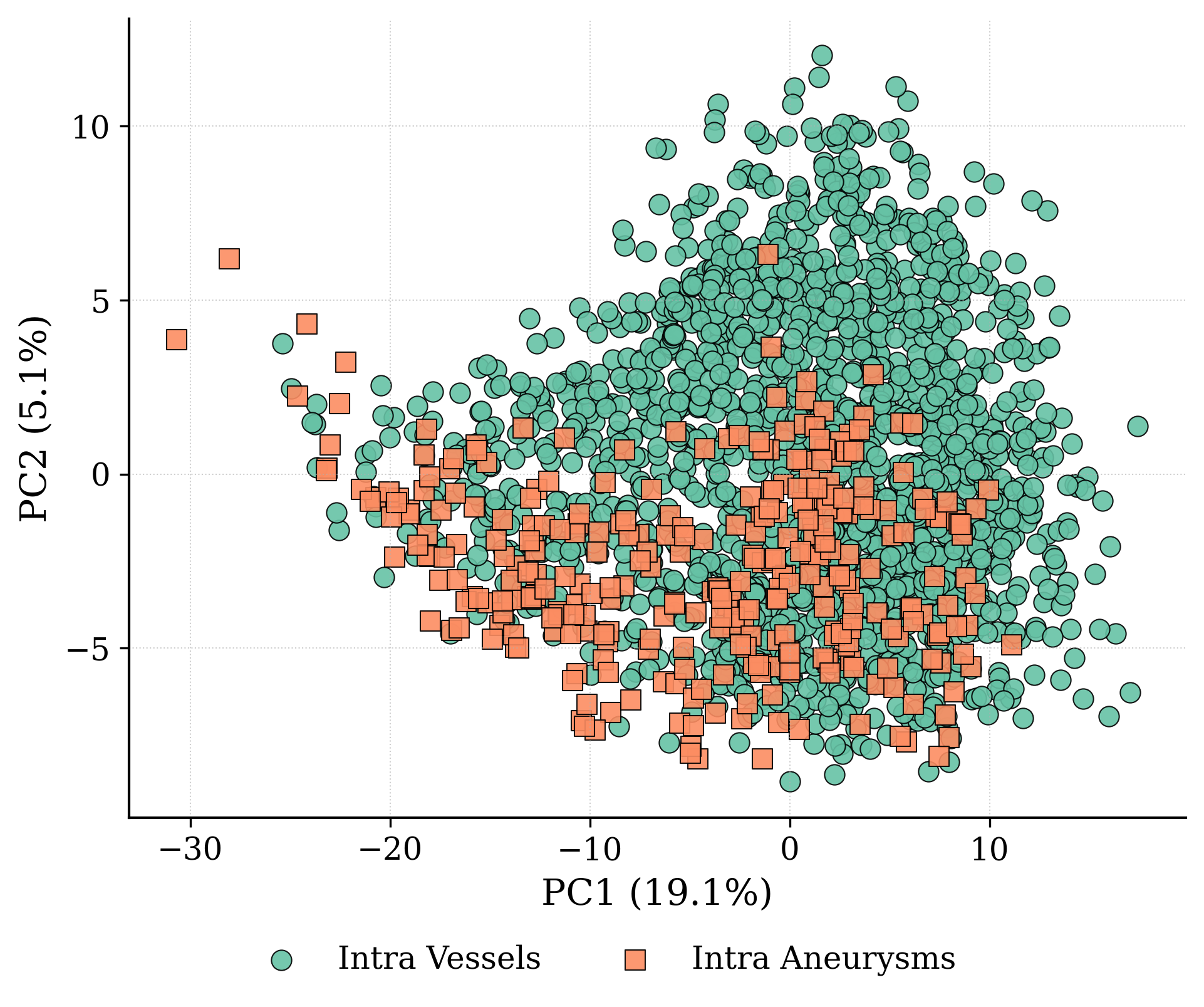}\\
    \includegraphics[width=0.32\textwidth]{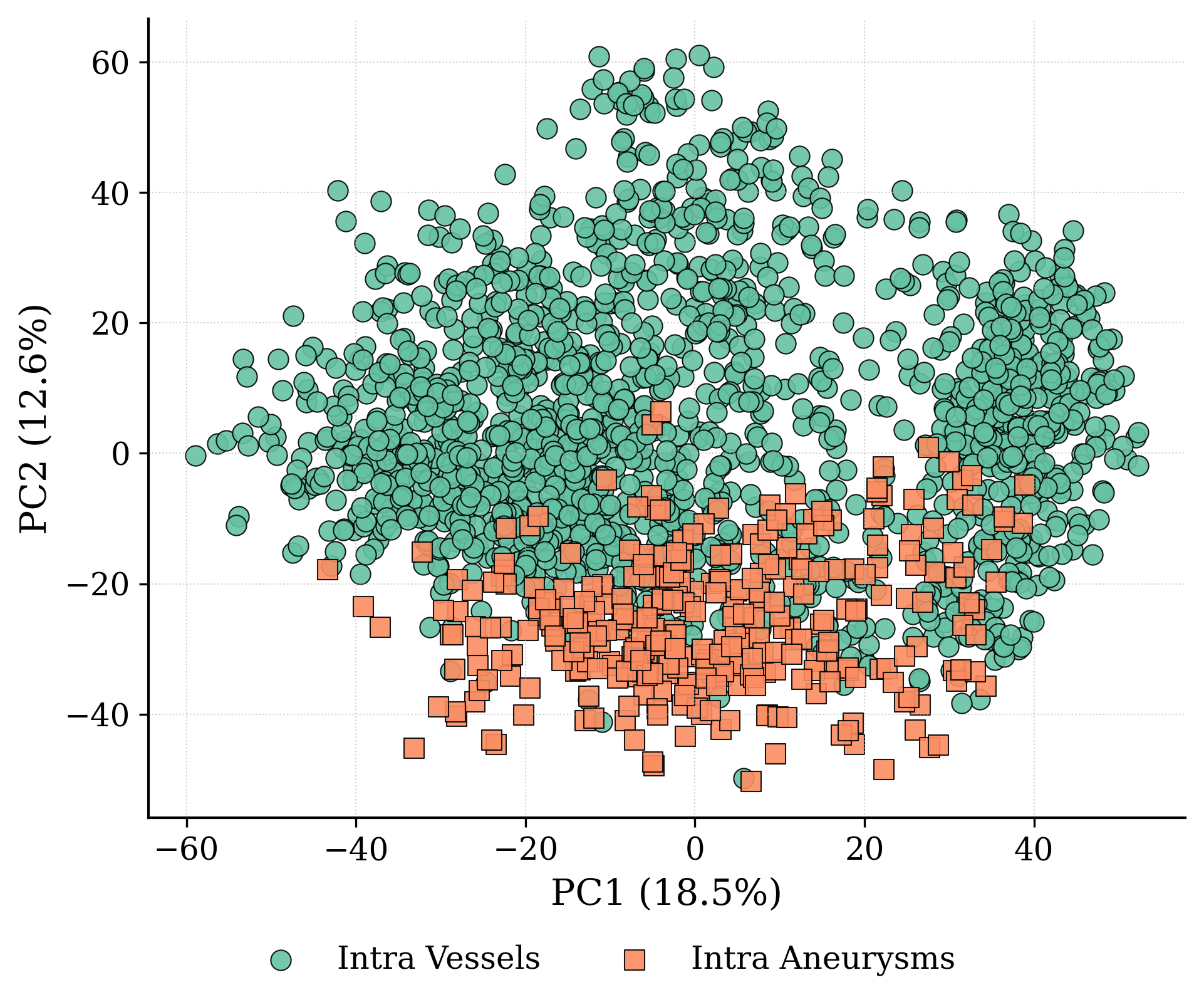}
    \includegraphics[width=0.32\textwidth]{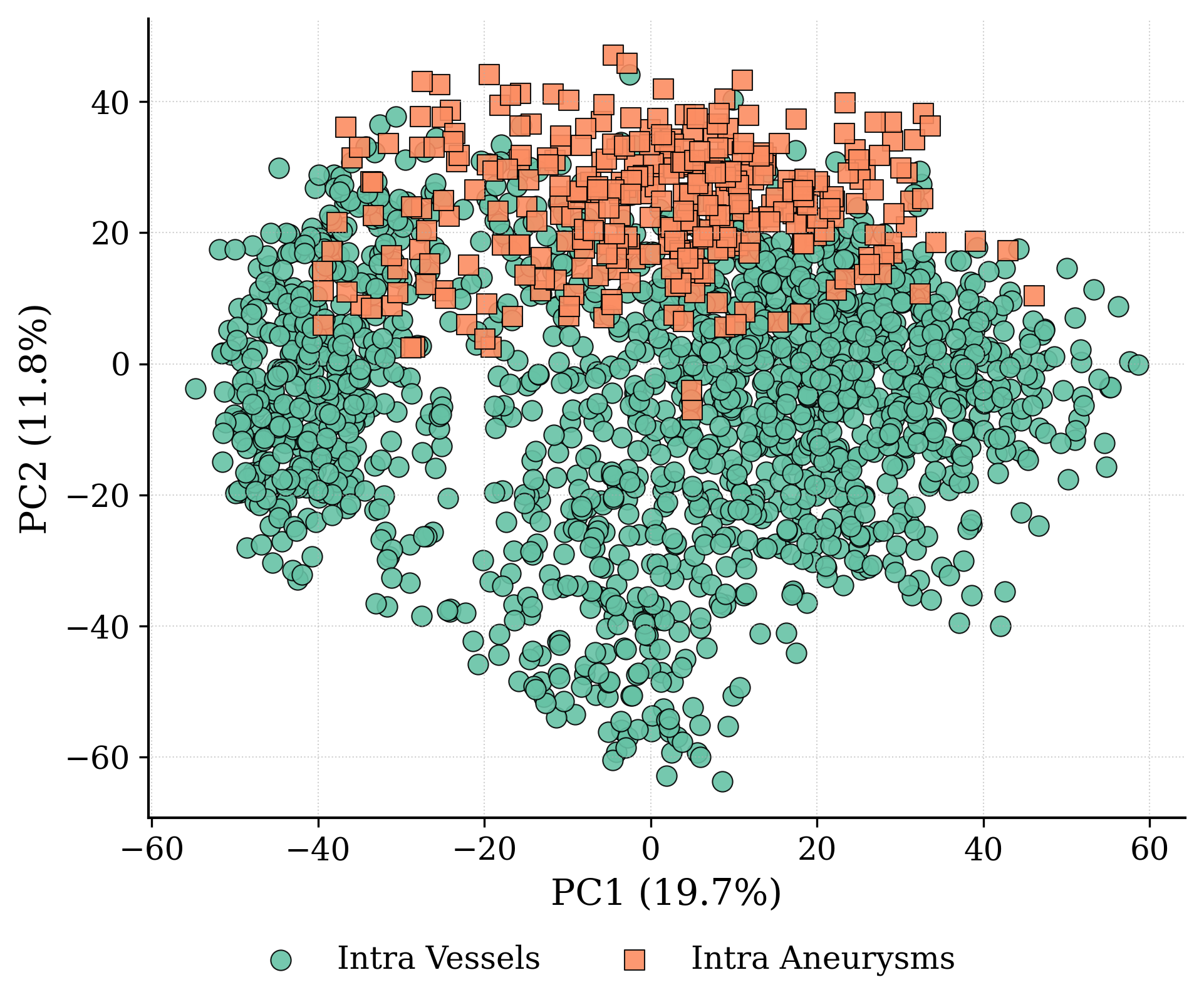}
    \caption{Results of PCA on the Intra3D dataset. Figures show the mean, standard deviation, minimum, and maximum of the PCA components.}
    \label{fig:pca_supplementary_intra}
\end{figure}

\begin{figure}[h!]
  \centering
  \includegraphics[width=0.34\textwidth]{t-sne_mean.png}
  \includegraphics[width=0.34\textwidth]{t-sne_std.png}
  \includegraphics[width=0.34\textwidth]{t-sne_min.png}
  \includegraphics[width=0.34\textwidth]{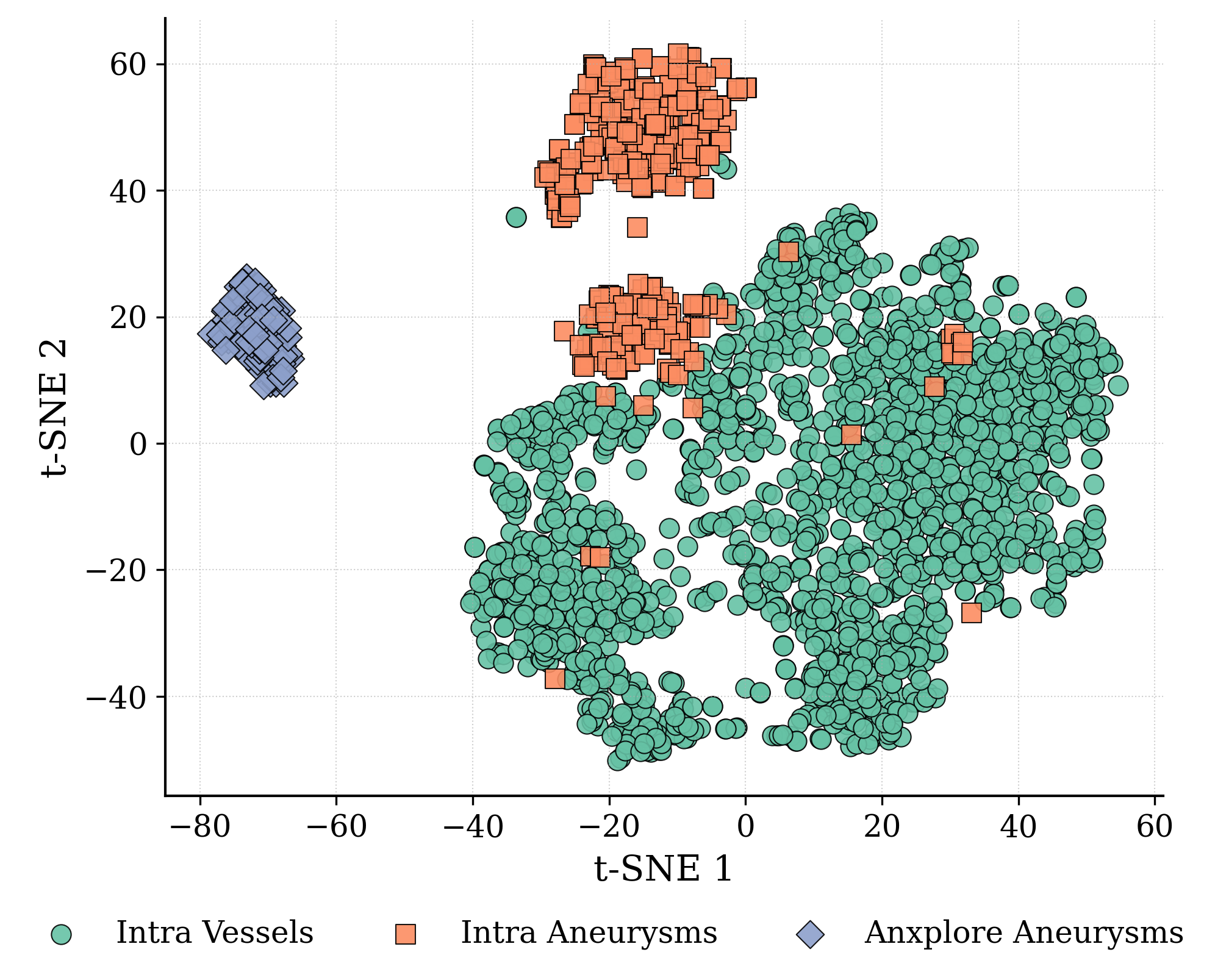}
  \caption{Result of t-SNE on the Intra3D dataset combined with the AnXplore dataset. Figures show the mean, standard deviation, minimum, and maximum of the t-SNE components.}
  \label{fig:tsne_supplementary}
\end{figure}

\begin{figure}[h]
  \centering
  \includegraphics[width=0.34\textwidth]{pca_intra_features.png}
  \includegraphics[width=0.34\textwidth]{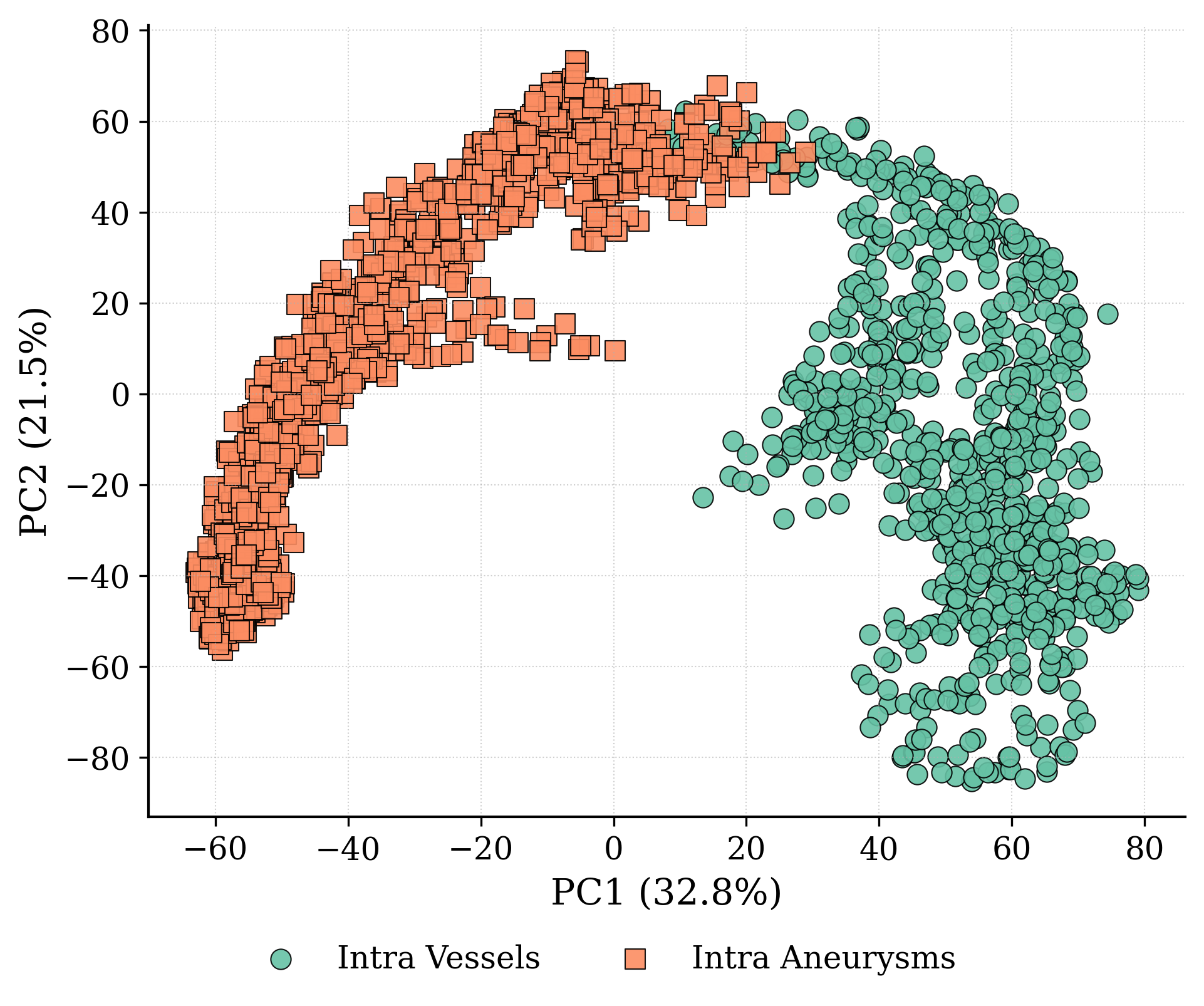}
  \includegraphics[width=0.34\textwidth]{t-sne_intra_features.png}
  \includegraphics[width=0.34\textwidth]{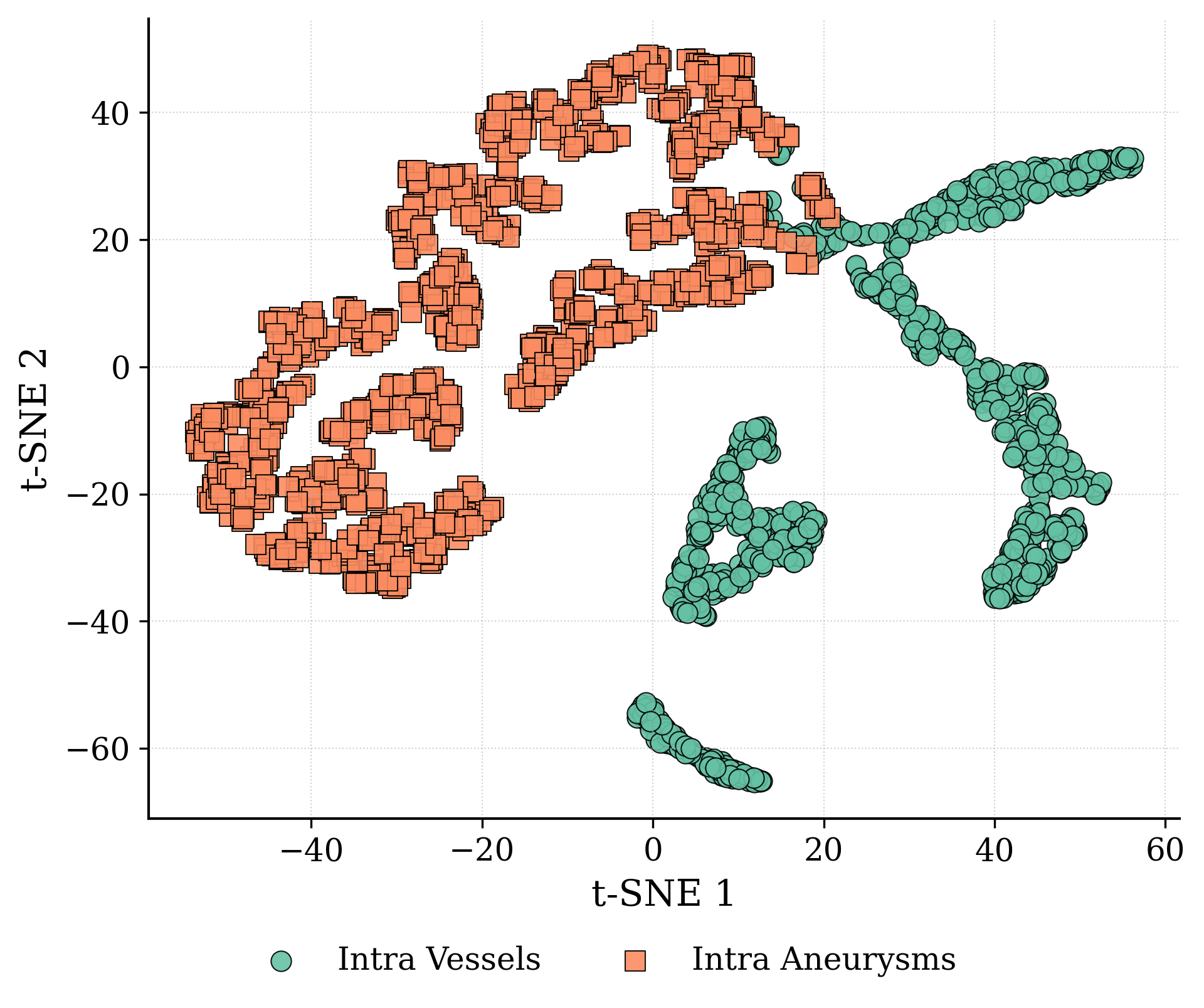}
  \caption{t-SNE and PCA on all points of two different aneurysms from the Intra3D segmentation dataset. The first two figures show the PCA components, while the last two figures show the t-SNE components.}
  \label{fig:seg_all}
\end{figure}

\begin{figure}[h!]
  \centering
  \includegraphics[width=0.34\textwidth]{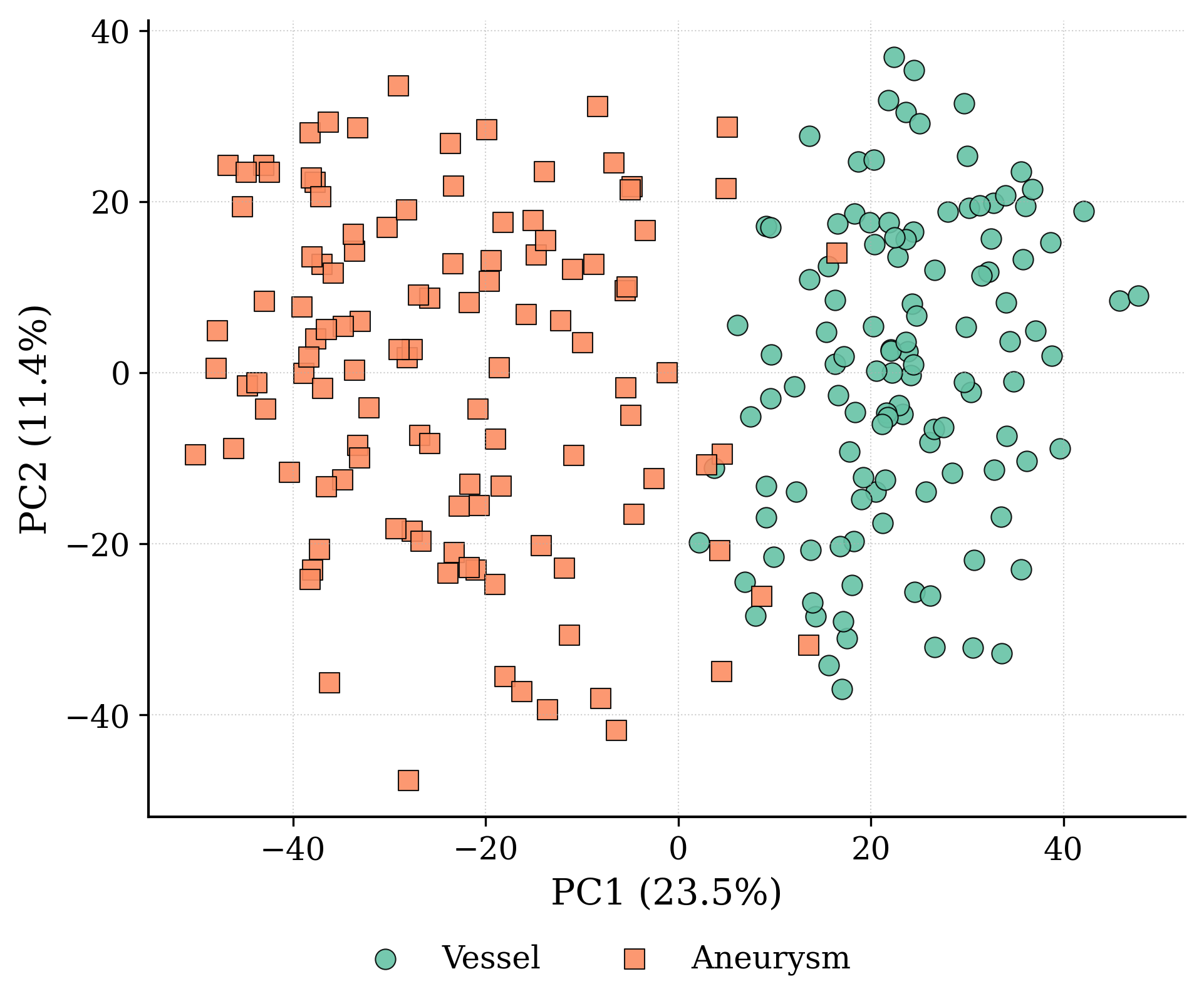}
  \includegraphics[width=0.34\textwidth]{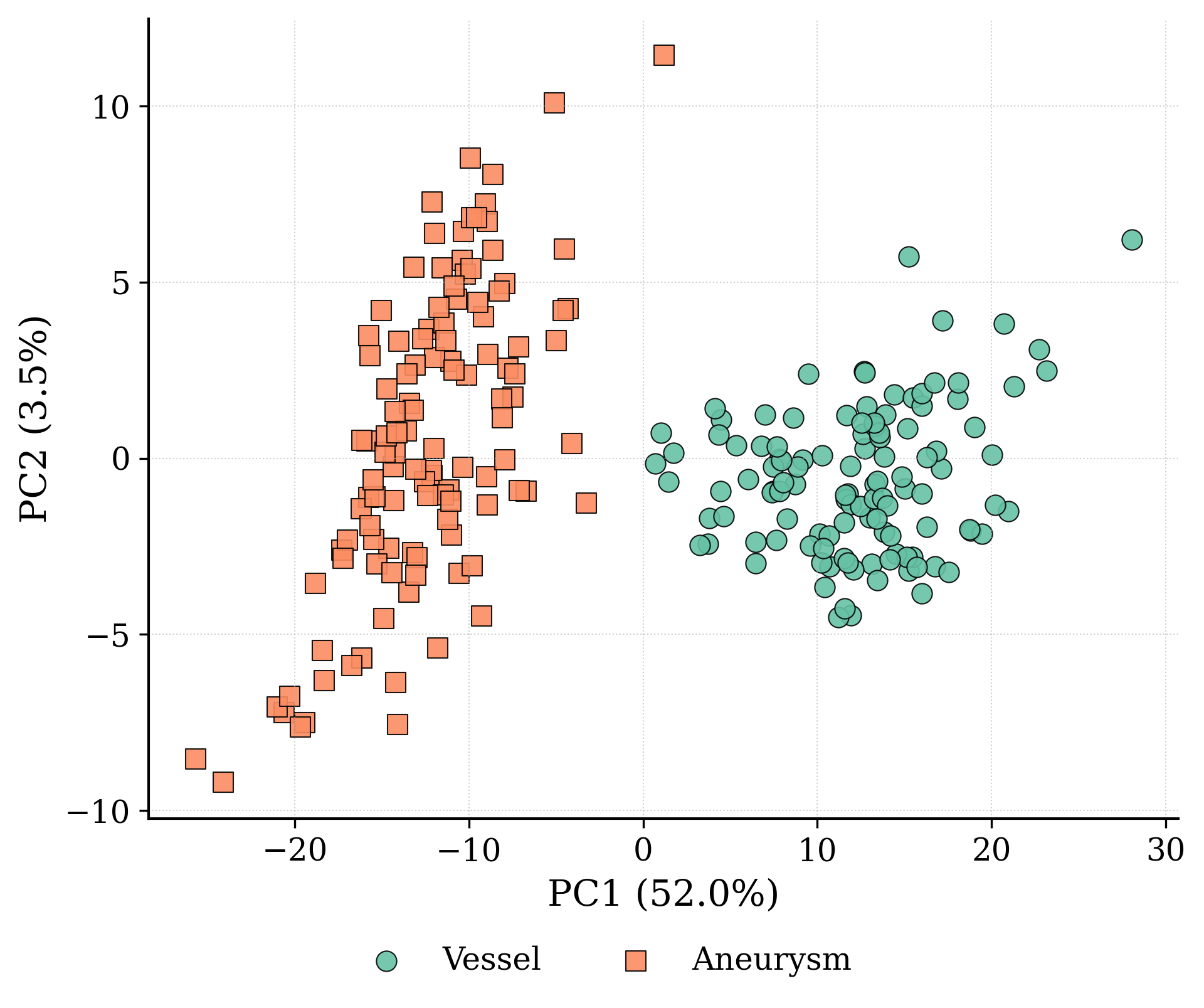}
  \includegraphics[width=0.34\textwidth]{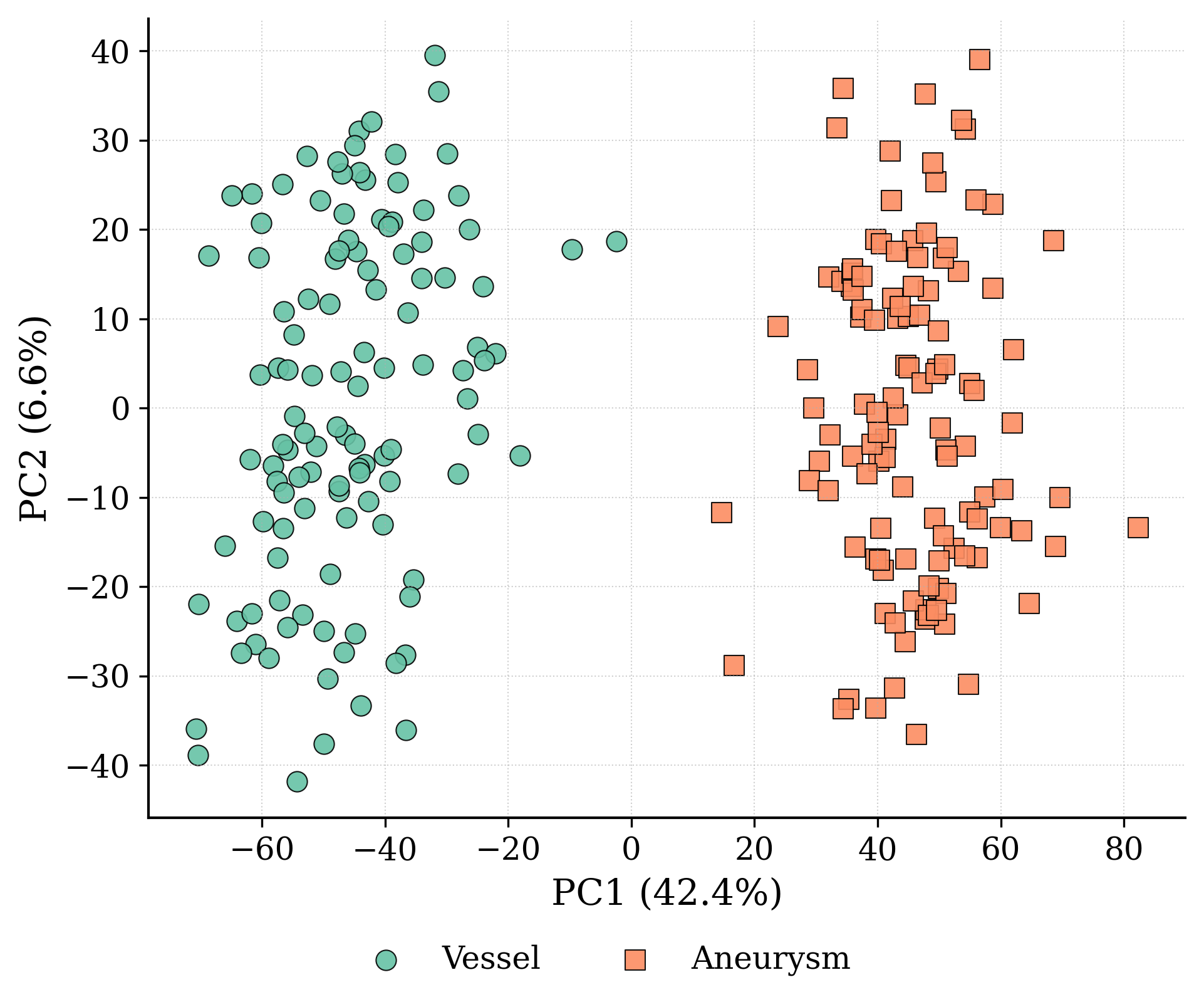}
  \includegraphics[width=0.34\textwidth]{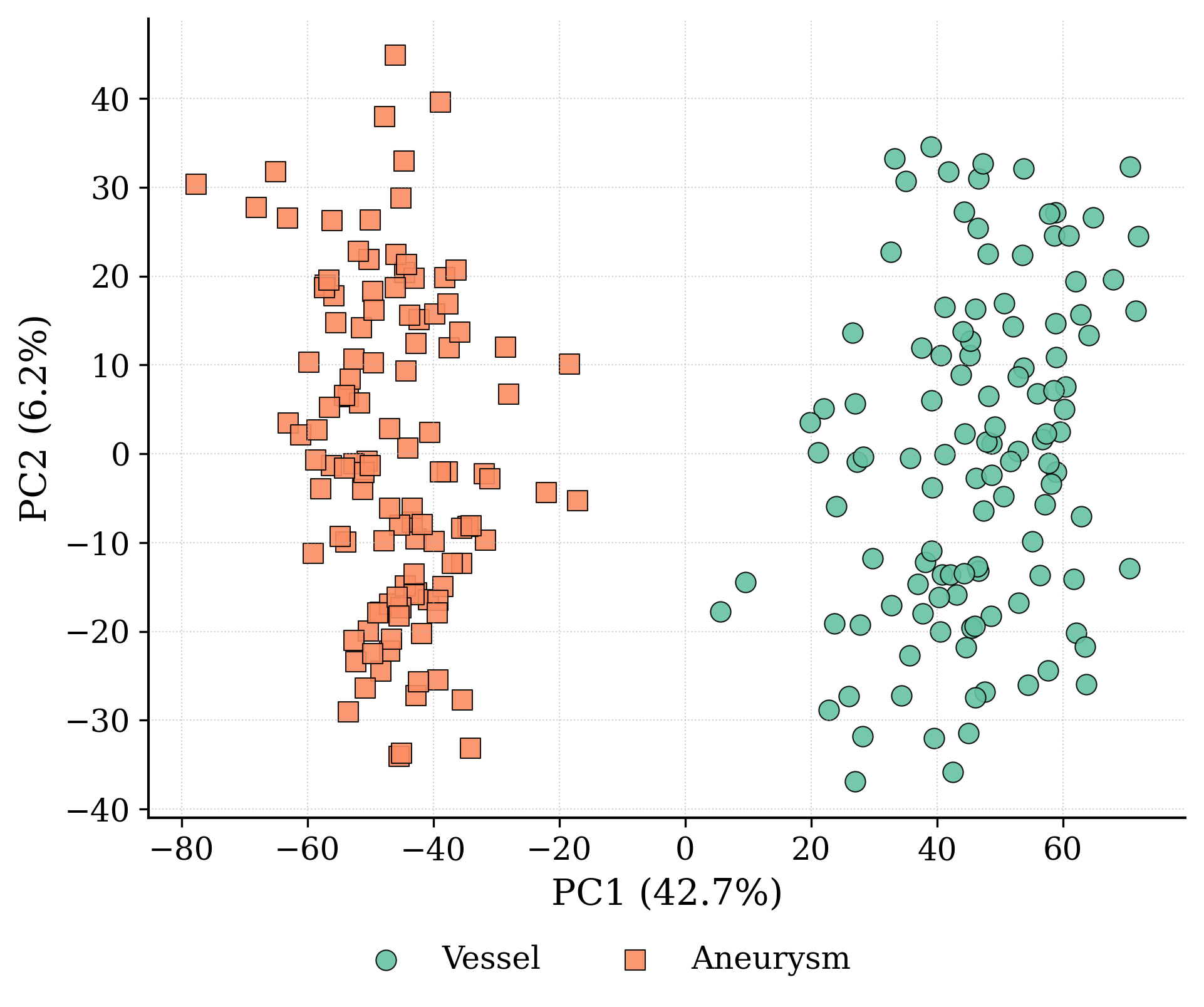}
  \caption{Results of PCA on annotated aneurysms over the Intra3D segmentation dataset. Figures show the mean, standard deviation, minimum, and maximum of the PCA components of the aneurysm and the vessel parts.}
  \label{fig:aneu_seg_supplementary}
\end{figure}

\begin{figure}[h!]
  \centering
  \includegraphics[width=0.34\textwidth]{t-sne_global_mean.png}
  \includegraphics[width=0.34\textwidth]{t-sne_global_std.png}
  \includegraphics[width=0.34\textwidth]{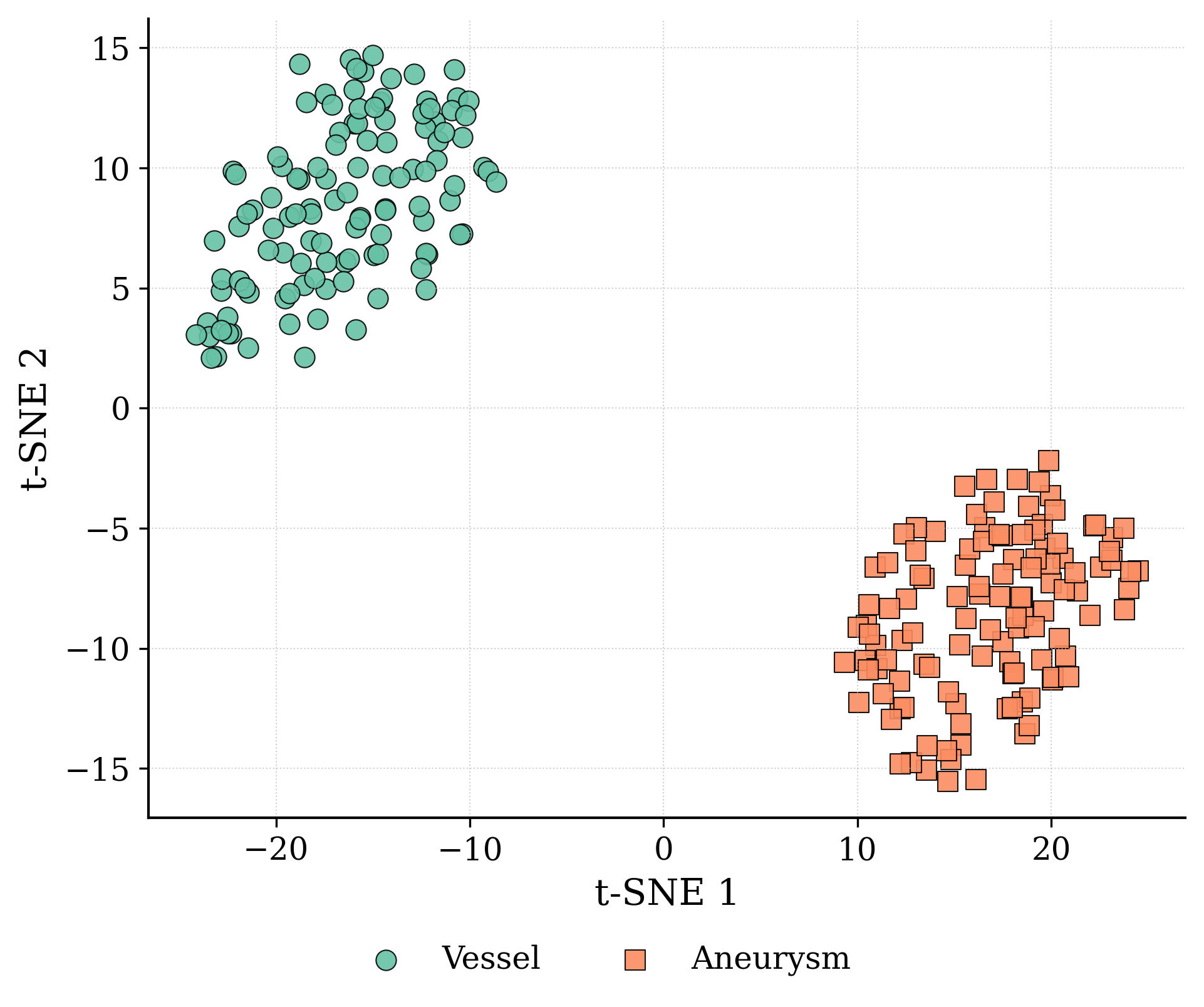}
  \includegraphics[width=0.34\textwidth]{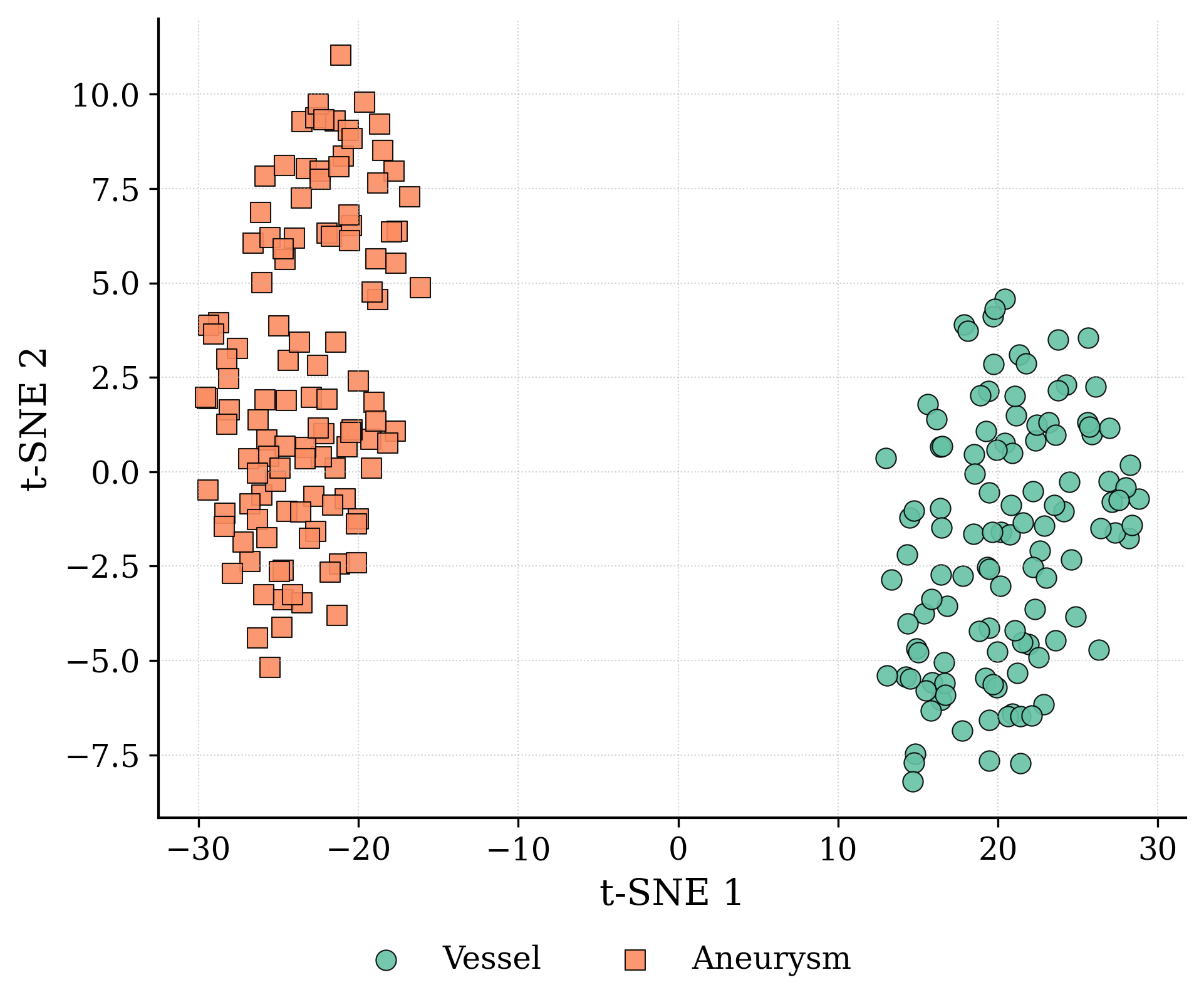}
  \caption{Results of t-SNE on annotated aneurysms over the Intra3D segmentation dataset. Figures show the mean, standard deviation, minimum, and maximum of the t-SNE components of the aneurysm and the vessel parts.}
  \label{fig:aneu_seg_supplementary_tsne}
\end{figure}

\begin{figure}[h!]
  \centering
\begin{center}
\resizebox{\textwidth}{!}{
\begin{tabular}{ccc}
    \includegraphics[width=0.330\textwidth]{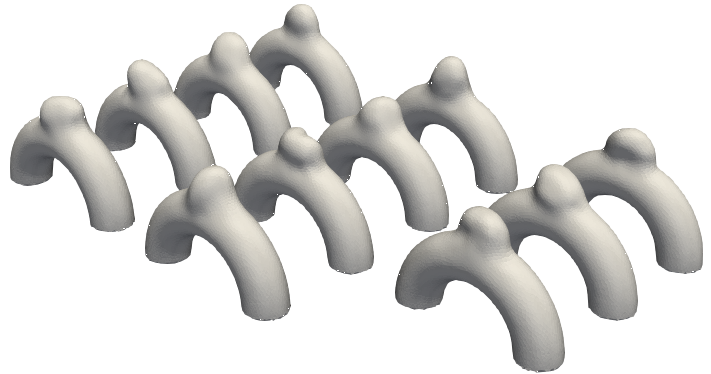} &
    \includegraphics[width=0.200\textwidth]{cluster_6.png} &
    \includegraphics[width=0.300\textwidth]{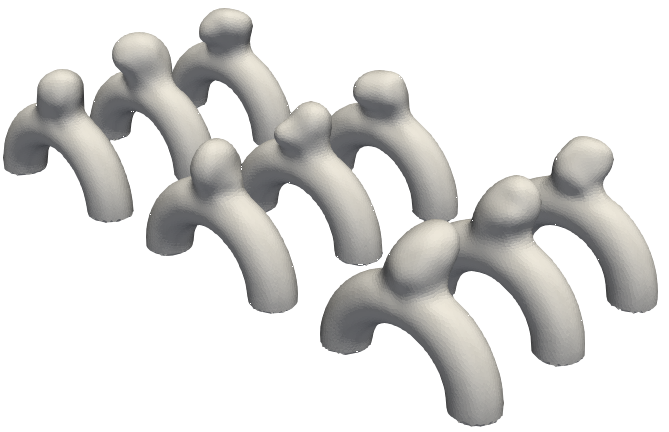} \\
    \includegraphics[width=0.249\textwidth]{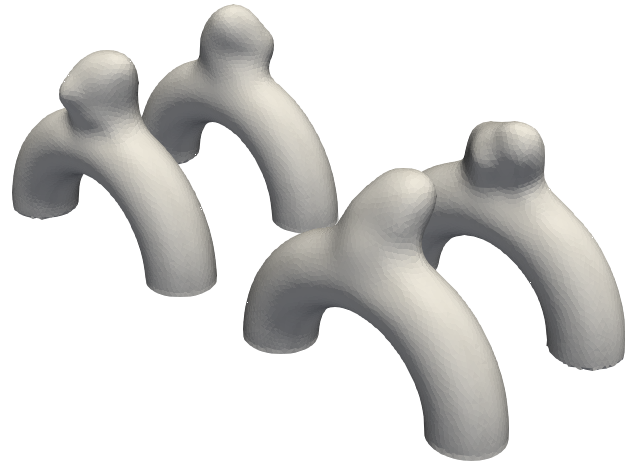} &
    \includegraphics[width=0.319\textwidth]{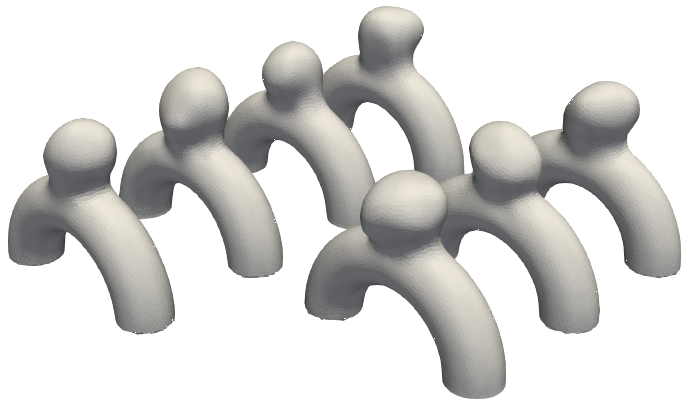} &
    \includegraphics[width=0.289\textwidth]{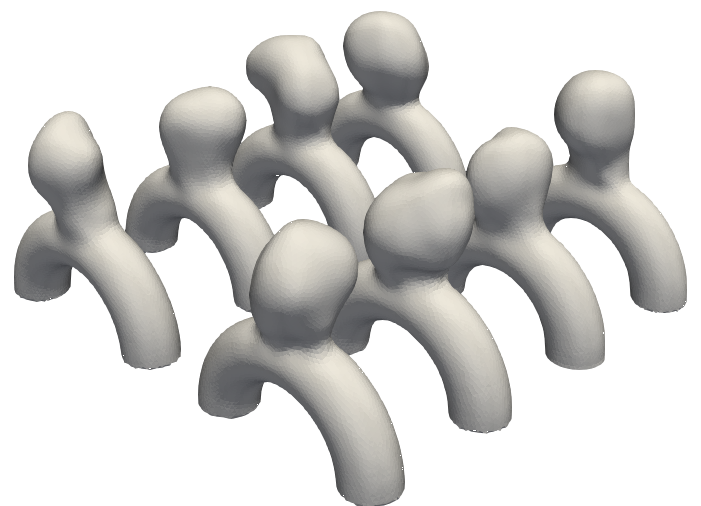} \\
    \includegraphics[width=0.330\textwidth]{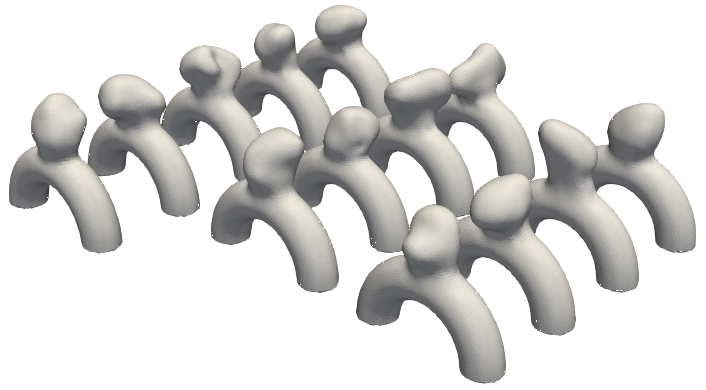} &
    \includegraphics[width=0.269\textwidth]{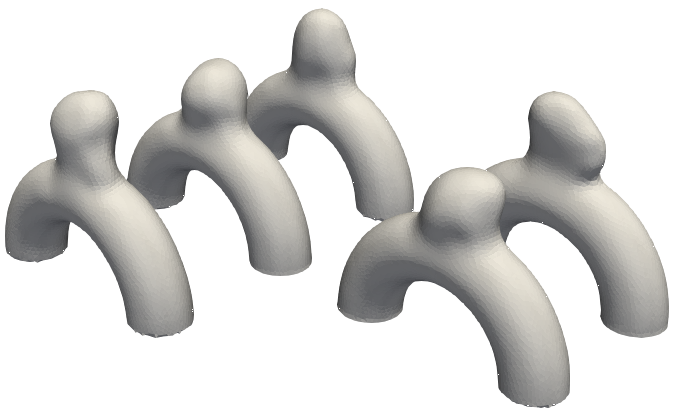} &
    \includegraphics[width=0.319\textwidth]{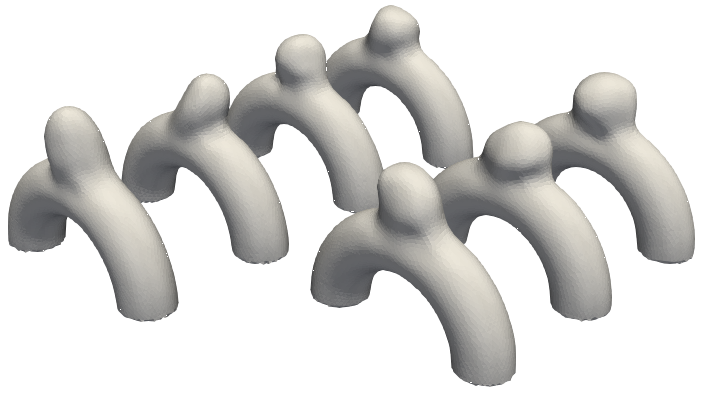} \\
    \includegraphics[width=0.310\textwidth]{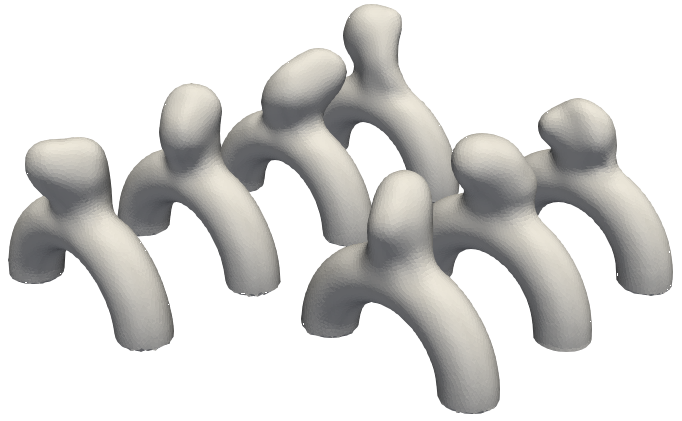} &
    \includegraphics[width=0.310\textwidth]{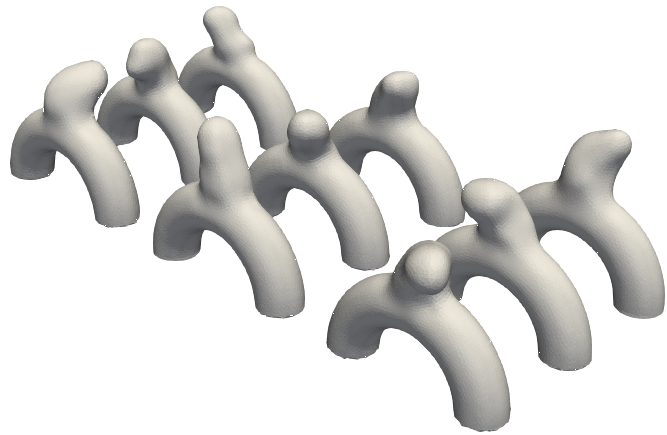} &
    \includegraphics[width=0.205\textwidth]{cluster_11.png} \\
    \includegraphics[width=0.314\textwidth]{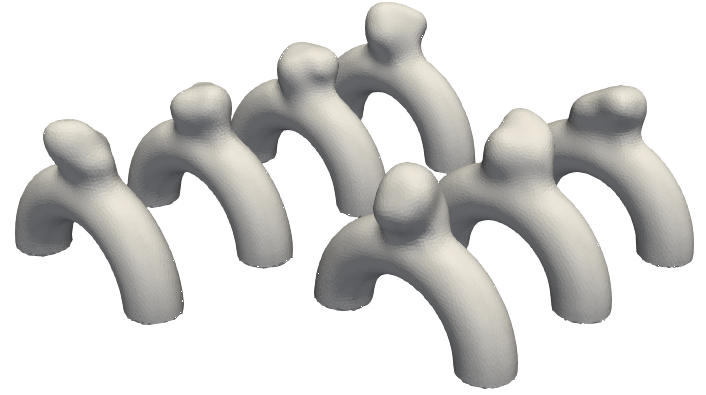} &
    \includegraphics[width=0.283\textwidth]{cluster_13.png} &
    \includegraphics[width=0.314\textwidth]{cluster_14.png} \\
\end{tabular}}
\end{center}
\caption{Clusters of aneurysms from the AnXplore dataset \citep{anxplore} based on the t-SNE from mean features of the aneurysms.}
  \label{fig:clustering_supplementary}
\end{figure}

\end{document}